\documentclass{article}

\usepackage{PRIMEarxiv}

\usepackage[utf8]{inputenc} % allow utf-8 input
\usepackage[T1]{fontenc}    % use 8-bit T1 fonts
\usepackage{hyperref}       % hyperlinks
\usepackage{url}            % simple URL typesetting
\usepackage{booktabs}       % professional-quality tables
\usepackage{amsfonts}       % blackboard math symbols
\usepackage{nicefrac}       % compact symbols for 1/2, etc.
\usepackage{microtype}      % microtypography
\usepackage{lipsum}
\usepackage{fancyhdr}       % header
\usepackage{graphicx}       % graphics
\graphicspath{{media/}}     % organize your images and other figures under media/ folder
\usepackage{csquotes}
\usepackage{graphicx}
\usepackage{multicol,caption}
\newenvironment{Figure}
  {\par\medskip\noindent\minipage{\linewidth}}
  {\endminipage\par\medskip}

\usepackage[dvipsnames]{xcolor}

\usepackage{dirtytalk}
\usepackage{hyperref}

\usepackage{amsmath,amssymb} 
\usepackage{subcaption}

\usepackage{multirow}
\usepackage{colortbl}
\usepackage{booktabs, chemformula}

\usepackage{tabularx}
\newcolumntype{Y}{>{\centering\arraybackslash}X}

\title{FacTeR-Check: Semi-automated fact-checking through Semantic Similarity and Natural Language Inference}

\author{
Alejandro Mart\'in \\
Universidad Politecnica de Madrid                                  \\
Madrid, Spain\\
\texttt{alejandro.martin@upm.es}      
\And
Javier Huertas-Tato \\
Universidad Politecnica de Madrid\\
Madrid, Spain\\
\texttt{javier.huertas.tato@upm.es} \\
\And

Álvaro Huertas-García \\
Universidad Politecnica de Madrid\\
Madrid, Spain\\
Universidad Rey Juan Carlos de Madrid\\
Madrid, Spain\\
\texttt{alvaro.huertas@upm.es} \\
\And
Guillermo Villar-Rodríguez \\
Universidad Politecnica de Madrid\\
Madrid, Spain\\
\texttt{guillermo.villar@upm.es} \\
\And 
David Camacho \\
Universidad Politecnica de Madrid\\
          Madrid, Spain\\
          \texttt{david.camacho@upm.es}}

\begin{document}

%\author{Alejandro Martín,
%        Javier Huertas-Tato,
%        Álvaro Huertas-García,
%        Guillermo Villar-Rodríguez,
%        David Camacho% <-this % stops a space
%\thanks{Alejandro Martín, Javier Huertas-Tato, Guillermo Villar-Rodríguez and David Camacho were with Department
%of Computer System Engineering, Universidad Politécnica de Madrid, Spain, e-mail: alejandro.martin@upm.es,  javier.huertas.tato@upm.es, guillermo.villar@upm.es, david.camacho@upm.es}% <-this % stops a space
%\thanks{Álvaro Huertas-García was with the Department of Computer Science, Universidad Rey Juan Carlos, Spain}% <-this % stops a space


\maketitle

\begin{abstract} Our society produces and shares overwhelming amounts of information through Online Social Networks (OSNs). Within this environment, misinformation and disinformation have proliferated, becoming a public safety concern in most countries. Allowing the public and professionals to efficiently find reliable evidences about the factual veracity of a claim is a crucial step to mitigate this harmful spread. To this end, we propose \textit{FacTeR-Check}, a multilingual architecture for semi-automated fact-checking that can be used for either applications designed for the general public and by fact-checking organisations. FacTeR-Check enables retrieving fact-checked information, unchecked claims verification and tracking dangerous information over social media. This architectures involves several modules developed to evaluate semantic similarity, to calculate natural language inference and to retrieve information from Online Social Networks. The union of all these components builds a semi-automated fact-checking tool able of verifying new claims, to extract related evidence, and to track the evolution of a hoax on a OSN. While individual modules are validated on related benchmarks (mainly MSTS and SICK), the complete architecture is validated using a new dataset called NLI19-SP that is publicly released with COVID-19 related hoaxes and tweets from Spanish social media. Our results show state-of-the-art performance on the individual benchmarks, as well as producing a useful analysis of the evolution over time of 61 different hoaxes.
\end{abstract}


%\IEEEpeerreviewmaketitle

\section{Introduction}
Misinformation and disinformation are two terms that resound since a long time. Inaccurate information has been largely used for varied purposes for decades and centuries. However, the emergence of Internet, Online Social Networks and Instant Messaging Services has undoubtedly facilitated its rapid creation and diffusion. These two terms reflect a problem that continues to expand and which involves an increasing concern to society. Yet, there are important differences between both terms: while misinformation involves inaccurate information propagated without knowing it is false, disinformation involves disseminating deliberately false information in order to deceive people\footnote{\url{https://dictionary.cambridge.org/es-LA/dictionary/english/disinformation}}. 

The COVID-19 pandemic has undoubtedly drawn attention to this problem, when misinformation and disinformation meet health and affect public safety. From the initiation of this pandemic, an incessant repetition of falsehoods has been generated and propagated, undermining the work of health authorities in the fight against COVID-19. False reports about its origin, its death rate, or about vaccines have been a constant threat to control this virus.

Fact-checking organisations are on the forefront combating the propagation of false claims, where intensive work is done to deny hoaxes that circulate through different channels, such as Online Social Networks (OSNs), Instant Messaging Services or Mass Media. The verification process conducted by these companies is mostly carried out by hand, however, it is barely reflected in OSNs. Users of these platforms share fake information without even realising it is indeed a falsehood or deliberately posting false claims without further consequences.

\begin{figure*}[htpb] 
	\centering
	\includegraphics[width=1\textwidth]{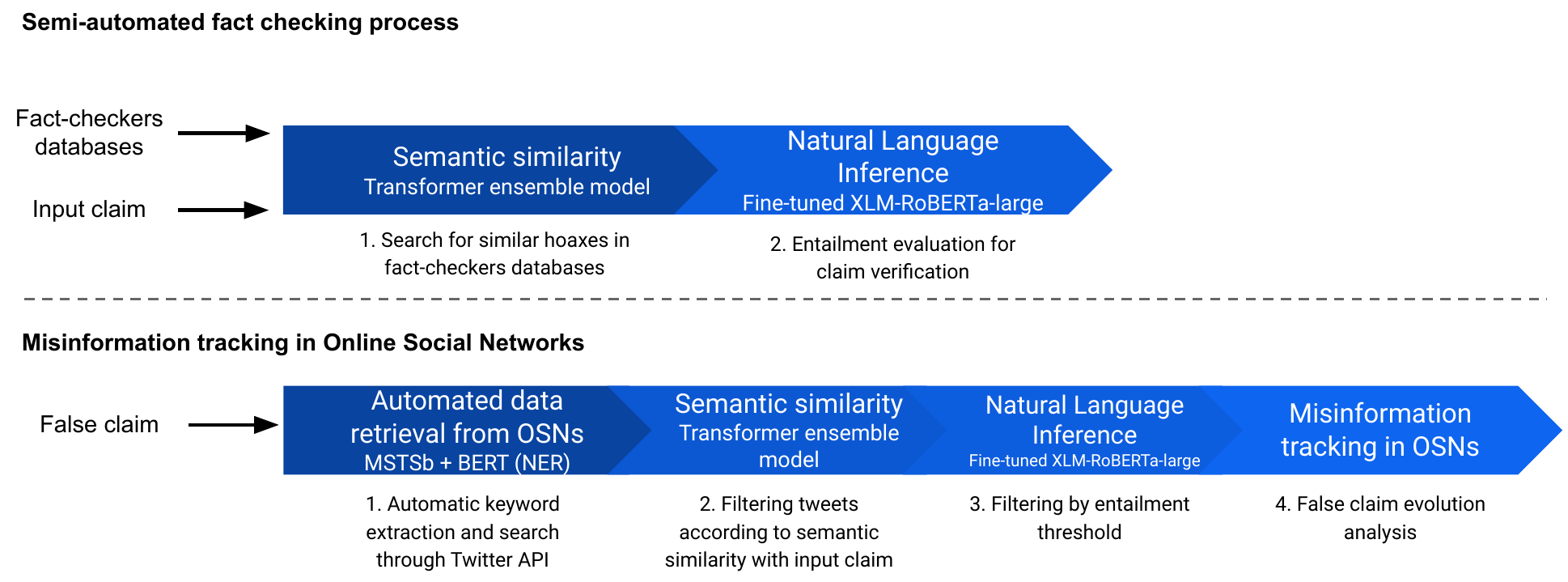}
	\caption{Diagram showing the two possible usage flows of FacTeR-Check.}
	\label{fig:facter_check}
\end{figure*}

Recent advances in Natural Language Processing, such as the Transformer architecture~\cite{vaswani2017attention}, allow to deal with complex human language for a plethora of tasks, such as summarization, translation, sequence classification, question answering or context-aware sentences similarity evaluation. The embeddings generated by this type of models for a piece of text, a vector representation composed of hundreds of 

In this research, we leverage the most recent advances in Natural Language Processing to develop a semantic-aware multilingual Transformer-based architecture for semantic similarity evaluation, semi-automated fact-checking and tracking of information pieces in Online Social Networks. We present an architecture that, on the one hand, can help general public in checking the veracity of a claim (i.e. a tweet) through context-aware automated comparison against a databases of hoaxes. On the other hand, our proposal aims at providing useful tools for fact-checking organisations to track and monitor hoaxes circulating in OSNs.

In contrast to previous approaches previously proposed, our tool relies on a semi-automated fact-checking process, using fact-checkers databases as source of verified claims. This ensures the quality of the predictions of the model, instead of relying on training sets of false data that severely limit the capacity of the model to detect the most recent. Another major difference lies in the context-aware and multilingual capacities we introduce due to the use of the Transformer architecture, a very important advance to deal with human language understanding and to allow comparisons between different languages without translation. The multilingual capacity will help to do fact check no matter the language of the candidate claim and the verified facts is. Finally, we also integrate a tracking module to analyse the whole propagation cascade of the hoax, a very valuable tool to explore its whole story in a social network.

To validate and to show the capabilities of the architecture proposed, we use the COVID-19 pandemic scenario in Spanish speaking countries. We manually selected 61 hoaxes related to Covid-19 and extracted related tweets using Twitter API. Our architecture allows to label the degree of entailment of these tweets with a hoax, providing a useful insight of the propagation of hoaxes in Spanish on Twitter throughout one year.

In summary, this research presents the following contributions: 

\begin{itemize}
    \item A labelled dataset of Spanish tweets IDs with a degree of entailment against a list of 61 hoaxes.
    \item A context-aware multilingual semantic similarity method for searching hoaxes with high similarity to a given query.
    \item A Natural Language Inference model for semi-automated fact-checking. This model allows to check if there is an entailment, contradiction or neutral relation between two statements.
    \item A deep insight of misinformation and disinformation circulating on Twitter related to Covid-19 in Spanish speaking countries during one year.
\end{itemize}

The remaining sections of this manuscript are organised as follows: Section~\ref{section:background} summarises a series of background concepts and the most relevant state-of-the-art works. Section~\ref{section:factercheck} presents the whole architecture designed for semi-automated fact-checking. Section~\ref{section:evaluation} reports the experiments conducted to evaluate the different modules that compose the FacTeR-Check architecture. Section~\ref{section:nli19-sp} presents the dataset built in this research of hoaxes found in Twitter and publicly released in this research. Section~\ref{section:analysis_twitter} provides a deep analysis of the propagation of hoaxes related to Covid-19 in Spanish in Twitter

%\cite{huertas2021sml}

 

\section{Background and related work}
\label{section:background}
In this section, a short selection of some relevant background work is presented together with an overview of the state-of-the-art literature. The section provides some recent contributions and works on transformers architectures (Sec.\ref{label:transformers}), semantic (textual-based) similarity methods (Sec.\ref{label:semantic}), natural language inference tasks (Sec.\ref{label:nli}), automated fact-checking (Sec.\ref{label:afc}), and misinformation tracking in OSN (Sec.\ref{label:misinformation}).

\subsection{The Transformer architecture}
\label{label:transformers}

In 2017, a group of researchers working at Google presented the Transformer~\cite{vaswani2017attention}, a novel network architecture based on the concept of \textit{attention} to deal with complex tasks involving human language, such as translation. This architecture revolutionised the Natural Language Processing field, allowing to train models to address highly complex tasks efficiently. From then, an uncounted number of applications, architectures, and models have been published to address tasks such as sentiment analysis~\cite{naseem2020transformer}, text generation~\cite{zhang2019bertscore} or question answering~\cite{yang2019end}. However, the attention concept was also soon exported to other domains such as music generation~\cite{zhang2020learning} or image generation~\cite{parmar2018image}.

One of the most important characteristics of these architectures in the Natural Language Understanding field lies in their context-aware capabilities, enabling to perform tasks such as question answering with high performance. While in previous NLP statistical-based approaches words were treated independently without considering the existing relations between them in a sentence or a text, the attention-based mechanism of the Transformer architecture allows to consider these relations and to establish deep connections. 

As in the case of other deep architectures such as Recurrent Neural Networks (RNNs) or Convolutional Neural Networks (CNNs), the Transformer involves a series of encoder and decoder layers that operate sequentially over the input. The goal of this architecture of this architecture is to obtain a vector representation called \textit{embedding} of the input sentence as comprehensive as possible to later be used in specific tasks. For instance, BERT is a specific implementation of the Transformer architecture where the output for a given input is an embedding of 768 positions that define multiple characteristics of the input. Due to the large amount of data, execution time and computational resources required to train this kind of models, researchers usually employ pre-trained architectures that are later fine-tuned to solve specific tasks. 

A plethora of architectures have been proposed implementing the attention-based mechanism since it was proposed. Models such as BERT~\cite{devlin2018bert}, Roberta~\cite{liu_roberta:_2019}, XML~\cite{conneau2019cross} or XLM-RoBERTa~\cite{conneau2019unsupervised} are being used in a large number of NLP tasks with great success.

\subsection{Semantic Textual Similarity}
\label{label:semantic}

Measuring the degree of similarity between a pair of texts is a problem that has attracted the attention of many researchers for many years from the natural language processing and information retrieval fields. The complexity of this task has resulted in a variety of approaches to obtain similarity measures able to consider the higher number possible of characteristics. Classical approaches relying on lexical based information have been largely used for this task, however, they are extremely limited, since they do not allow to compare the real semantic value~\cite{mihalcea2006corpus}. These methods fail to detect similarity between synonyms and they do not consider the existing relations between words of a sentence. Gomaa and Fahmy~\cite{gomaa2013survey} proposed a taxonomy of similarity methods. String-based similarity methods operate with string and characters sequences or ngrams~\cite{millar2000performance,singthongchai2013method}. Corpus-based methods use large sets of words and texts and metrics such as latent semantic analysis~\cite{dennis2003introduction} or building terms vectors~\cite{shrestha2011corpus}. Knowledge-based methods allow to use the semantic content to provide more accurate comparisons, usually employing semantic networks~\cite{schuhmacher2014knowledge}. The fourth category is composed of hybrid solutions combining different methods~\cite{mihalcea2006corpus}.

The proposal of using an attention-based mechanism and its implementation into the Transformer architecture has meant a turning point. The embeddings generated with this type of architecture of a sentence or a text allow to build a rich multidimensional space where multiple characteristics are represented, including the semantic value. Once obtained the embedding vector of each document to be compared, a spatial distance such as cosine similarity can be used to measure the degree of similarity. Pre-trained models can be used for this purpose. However, if these models do not provide enough precision, they can be fine-tuned in a specific domain thus allowing more accurate similarity calculation. When training these models in a multilingual scenario, they generate a common features space for all languages represented in the training data, thus enabling to compare texts in different languages. This capability has revolutionised the Natural Language Processing research field. 

However, building precise models implies to narrow the application domain, specialising in a specific task but loosing generalisation ability. As an example, transformers such as BERT have been combined with topic models to better deal with domain-specific language~\cite{peinelt2020tbert}. Researchers have also identified limitations in the use of general purpose Transformers~\cite{kasnesis2021transformer} due to the computational resources required to generate an embedding for each sentence to be compared but also because these representation embeddings are of low quality. Sentence-oriented models such as Sentence-BERT~\cite{reimers_sentence-bert:_2019} provide better sentence embeddings through the use of siamese and triplet network architectures together with a pooling operation applied to the output of BERT or RoBERTa and the cosine similarity metric. Datasets such as STS benchmark~\cite{cer2017semeval} or SICK~\cite{Marelli_Menini_Baroni_Bentivogli_Bernardi_Zamparelli_2014} are usually employed to train and evaluate these models. 

\subsection{Natural Language Inference}
\label{label:nli}

Natural Language Inference (NLI) is a NLP task where the goal is to evaluate if a sentence called hypothesis can be inferred given a sentence called premise~\cite{maccartney2009natural}. In other terms, given two sentences $a$ and $b$, is possible to infer if there is \textit{entailment} between them, which means that $b$ is based on $a$, if there is a \textit{neutral} relation, where $b$ could be true based on $a$ or if the relation is a \textit{contraction}, meaning that $b$ is not true based on $a$~\cite{gururangan2018annotation}. In the three cases, the pair of sentences could involve high similarity, but detecting an entailment relation goes a step further, involving deeper natural language understanding models.

There are different datasets which have been designed to train and evaluate NLP models for NLI, however, they are also typically used to train general-purpose Transformers given the importance of this task in Natural Language Understanding. The Stanford Natural Language Inference (SNLI) corpus~\cite{bowman2015large} is a corpus with 570,000 pairs of sentences labelled with contradiction neutral or entailment by 5 human annotators. Multi-Genre Natural Language Inference (MultiNLI)~\cite{williams-etal-2018-broad} to overcome several limitations of the SNLI dataset, where all sentences are extracted from image captions. MultiNLI is presented as a more complex corpus with a more varied language. Cross-lingual Natural Language Inference corpus (XNLI)~\cite{conneau2018xnli} was built to serve as a cross-lingual corpus including sentence pairs from 15 different languages. Recurring neural networks have proved to be able to achieve high performance in this domain, as it is the case of Long short-term memory networks (LSTMs)~\cite{chen2017enhanced,conneau2017supervised}. A number of Transformer-based approaches have also been proposed, allowing to compare inter-lingual sentences~\cite{huertas2021sml}.

NLI plays a very important role in automated fact-checking. Given a collection of false claims, the verification of a new information piece can be modelled as a NLI task where our goal is to detect entailment with one of the false claims collected. Similarly, given a collection of true facts, we can model as a NLI task the process of determining if a new fact is true based on the existing facts in that collection.

\subsection{Automated Fact-Checking}
\label{label:afc}

Automated Fact-Checking (AFC) involves different tasks and issues, such as extracting check-worthy claims from a speech or a large text, building fact-checking tools based on previously checked facts or to evaluate at what level a claim can be considered true. These AFC methods typically integrate Machine Learning techniques, however, researches have also highlighted the limitations of these approaches due to the training set used or the detection of paraphrasing~\cite{graves2018understanding}. Nevertheless, recent advancements in this field, mainly because of the development of architectures using the attention-based mechanism, have led to important progress in the area.

Typically, Automated Fact-Checking is usually conducted through NLP models. There are different approaches to address this task according to the inputs~\cite{thorne2018automated}. One possibility is to derive the veracity of a claim without further knowledge or context~\cite{granik2017fake}, an approach highly unreliable. Similarly, a multi-source approach has been proposed to combine different information sources~\cite{karimi2018multi}. Other researchers leverage knowledge to reach more reliable decisions. FEVER is a dataset of claims extracted from Wikipedia and manually labelled as \textit{Supported}, \textit{Refuted} or \textit{NotEnoughInfo}~\cite{thorne2018fever}. Hanselowski et al.~\cite{hanselowski2019richly} made public another dataset for automated fact-checking, with validated claims and documents annotated. WikiFactCheck-English~\cite{sathe2020automated} contains claims, context information and evidence documents. A comparative transformer-based approaches for misinformation detection is presented by Huertas et al.~\cite{huertas2021civic}.

These datasets are usually employed to train machine learning-based tools for AFC to later classify news claims without considering recent knowledge~\cite{miranda2019automated}. From another point of view, literature can also be organised according to how technology helps fact-checkers. An analysis study by Nakov et al.~\cite{nakov2021automated} identifies several tasks: searching for check-worthy claims, identifying already fact-checked claims, searching for evidences or providing automated fact-checking services.

In terms of specific implementations for AFC, Naderi and Hirst~\cite{naderi2018automated} use linguistic features and a classifier in a statement multi-class classification task. Karadzhov et al. propose the use of LSTM networks to classify claims in combination with relevant fragments of text from external sources~\cite{karadzhov2017fully}. Kotonya et al.~\cite{kotonya2020explainable} provide a broad analysis of the state-of-the-art literature of automated fact-checking approaches that are focused on explainability. Other important implementation is ClaimBuster~\cite{hassan2017toward}, which monitors live discourses and detects claims that are present in a repository, however limited details are provided regarding its implementation and there is no mention to the use of context-aware semantic models. More recent approaches have made use of the Transformer architecture. Stammbach and Ash~\cite{stammbach2020fever} use GPT-3 to generate a summary of evidences for a fact check decision. The attention-based mechanism has been also used for the identification of check-worthy statements~\cite{pathak2021self}. BERT has been also used for veracity prediction and explanation generation in a public health domain~\cite{kotonya2020explainable}.

\subsection{Misinformation tracking in OSNs}
\label{label:misinformation}

Online Social Networks (OSNs) are the perfect environment for a fast and uncontrolled growth of misinformation and disinformation. The effects produced by the complex opinion dynamics that occur in these platforms such as \textit{polarisation}, \textit{echo-chambers}, \textit{peer presure} or \textit{social influence}~\cite{ruffo2021surveying} hinder the process of analysing the propagation of a false claim. Monti et al.~\cite{monti2019fake} propose the use of Geometric Deep Learning to detect false claims in Online Social Networks, an approach which allows to take into consideration the propagation as a graph. A similar approach is followed by FakeDetector~\cite{zhang2020fakedetector}, in this case using a graph neural network and explicit and latent features to represent both text, creators and subjects. With a different objective, researchers have propose the use of transformers for profiling hate speech in Twitter~\cite{huertas2021profiling}.

The fight against misinformation in Online Social Networks has also been explored from an author perspective, modelling user profiles and their characteristics according to the probability to trust or distrust false claims~\cite{shu2018understanding,shu2019role}.

\section{Fighting misinformation through Semantic Similarity and Natural Language Inference}
\label{section:factercheck}
    
    %This section describes our proposal to fight against misinformation and disinformation disseminated through Online Social Networks. A general overview of the architecture is presented first, followed by a deep description of each functionality. 
    
    FacTeR-Check aims at helping in the whole verification process, analysis and tracking of false claims mainly circulating on social networks. Our tool implements an interconnected architecture with multilingual and deep human language understanding capabilities, substantially differing from previous completely automated but limited methods proposed in the literature relying on an initial immutable knowledge base. These methods used to train a machine learning classifier which fail when zero-shot prediction is performed, that is to say, when a claim which has never been verified by fact-checkers is presented. Instead, given the undeniable need to provide answers based on updated information sources, FacTeR-Check leverages the work already being conducted by fact-checking organisations to validate new claims. This semi-automated fact-checking process implies a close joint working between computational intelligence experts and fact-checking organisations.

    \begin{figure*}[htpb] 
    	\centering
    	\includegraphics[width=1\textwidth]{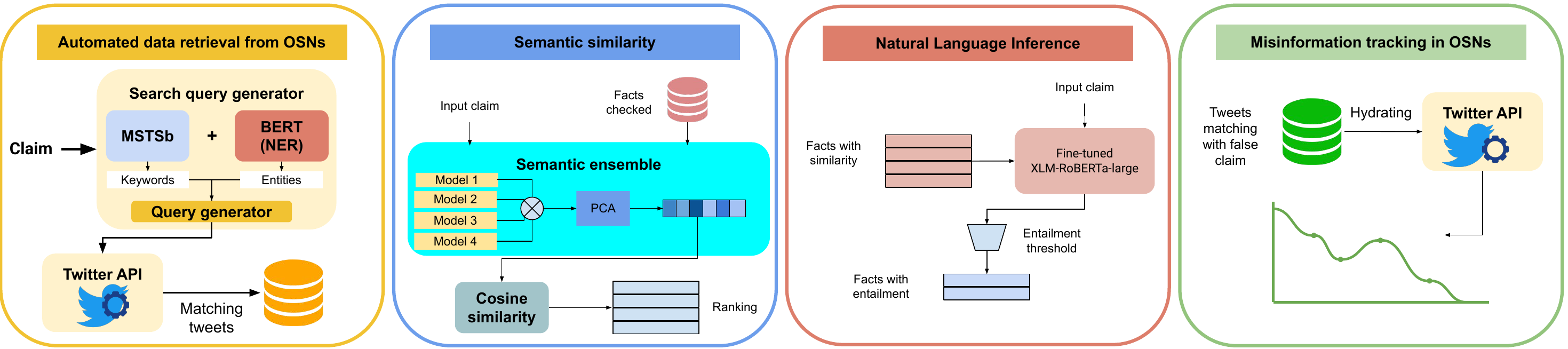}
    	\caption{The four components composing the FacTeR-Check architecture.}
    	\label{fig:boxes}
    \end{figure*}

    Besides, FacTeR-Check not only helps during the fact-checking process, but also in the collection and analysis of the whole history of a hoax, automatising the process of obtaining a broad oversight of its propagation over time. This is a powerful instrument to fight against mis- and disinformation spreading on social networks. FacTeR-Check provides four different main functionalities (see Fig.~\ref{fig:boxes}):

    \begin{enumerate}
    
        \item \textbf{Multilingual Semantic similarity evaluation:} For each new claim received, the architecture searches for semantically-similar hoaxes verified by fact-checkers in a database constantly updated. We make use of an ensemble of Transformer models to generate a representation embedding for each claim present in the database and for the one received as input. Then, a similarity distance is used to calculate the most similar hoaxes.
        
        \item \textbf{Multilingual Natural Language Inference:} Once a selection of similar hoaxes is presented, a NLI modules calculates the entailment probability with the input claim. If a coincidence is found (an entailment probability exceeds a certain threshold), the input claim is consider as false information. This module also allows to detect if the input claim denies or contradicts the hoax.
        
        \item \textbf{OSN automated retrieval:} In order to study the level of spread and presence of the hoax on a particular Online Social Network, a query containing a series of relevant keywords is created and send it to the API of the OSN. This enables to collect posts or tweets of users related to a false claim to be tracked. This step includes two transformer-based models for keyword extraction and Named Entity Recognition.

        \item \textbf{Misinformation tracking in OSNs:} Based on the three previous functionalities, it is possible to extract a pool of claims from OSNs and to filter those which replicate and support a false claim used as input. This module allows to analyse a large set of posts or tweets according to their creation date, user or other metadata.

    \end{enumerate}

\begin{figure*}[htpb] 
    	\centering
    	\includegraphics[width=0.9\textwidth]{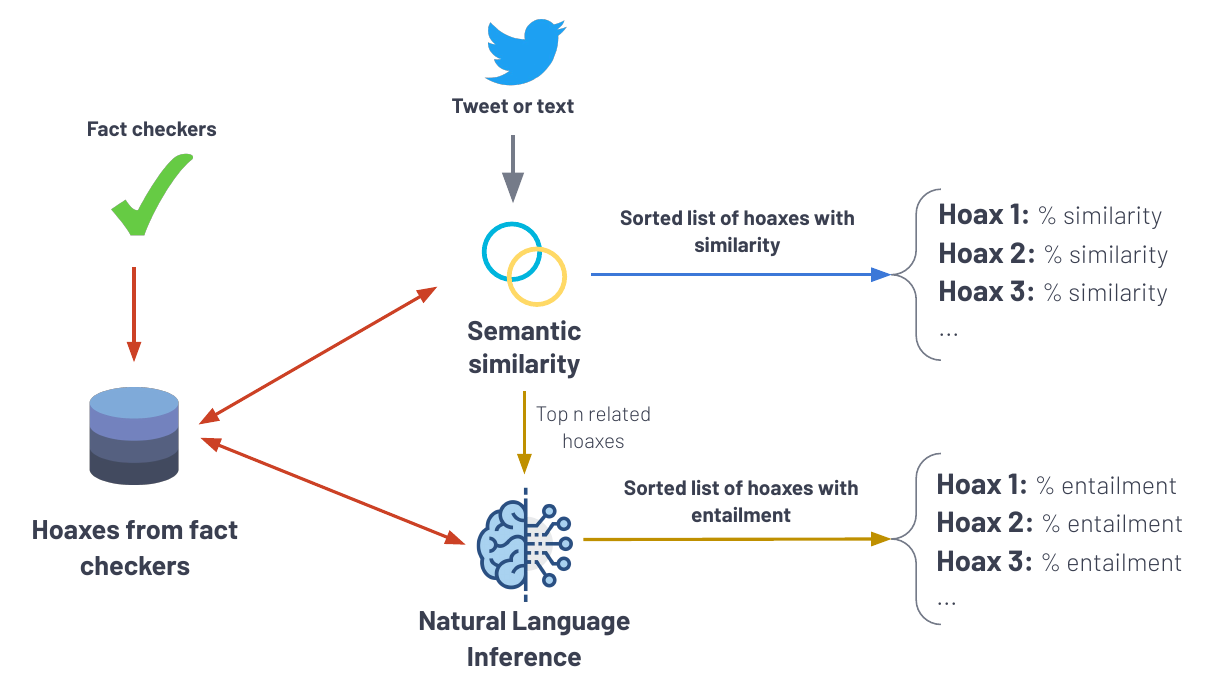}
    	\caption{Architecture for the evaluation of information pieces against hoaxes already identified by fact checkers. A first step allows to retrieve hoaxes that are semantically similar to the input text. In the second step, a Natural Language Inference model measures the degree of entailment against each hoax retrieved in step 1.}
    	\label{fig:1_topology}
    \end{figure*}

    The four functionalities described enable two different workflows, as shown in Fig.~\ref{fig:1_topology}. One is intended to provide a useful mechanism for a semi-automated fact verification, checking claims against a database of facts verified by fact-checking organisations. This workflow requires a semantic similarity module for filtering facts according to a certain degree of similarity and a second step of Natural Language Inference, to detect if there is textual entailment. 
    
    The second workflow is designed to aid them in the process of monitoring and tracking the life of a false claim in an Online Social Network. This involves to extract relevant keywords and named entities from the claim to build a search query which is sent to the API of the OSN in order to extract tweets or posts presenting content related to the input claim. The semantic similarity and NLI modules allow then to filter all the data to keep tweets or claims actually supporting the false claim. Next subsections describe in detail each functionality.

    %Previous approaches in the state-of-the-art literature have focused on a specific task to address the publication and dissemination of false claims in Online Social Networks, but they have failed in addressing the problem 

    \subsection{Semantic Similarity}
        \label{section:semantic_similarity_description}
        
        % Semantic similarity 
        Semantic is the level of language that deals with the meaning of a sentence by focusing on word-level interactions. This research aims to infer and understand information from texts in order to tackle misinformation by comparing sentence embeddings that condense the semantic level of language. In contrast to previous approaches focused on statistical natural language processing, FacTeR-Check implements semantic and context-aware semantic similarity evaluation. Through the use of semantic-aware and context-aware models, the goal is to evaluate the degree of similarity between a new claim against and a database of fact-checked claims. The result will be a subset of fact-checked claims ensuring a certain minimum degree of similarity.
        
        % Cosine similarity
        To measure the semantic similarity between texts, the cosine similarity function can be used. This metric takes advantage of the text representation as a vector in a high-dimensional space to compute the semantic and contextual proximity between a pair of texts, an operation which enables to assess their semantic similarity. The cosine distance between two sentence embeddings \textit{u} and \textit{v} is a variant of the inner product of the vectors normalised by the vectors' L2 norms, as shown in equation~\ref{eq_1}:
        
        % Insertar equation
        \begin{equation}\label{eq_1}
        CosSim(u,v) = \dfrac{ \sum_{i=1}^{N} u_i v_i }{\sqrt{\sum_{i=1}^{N} u_i^2} \sqrt{\sum_{i=1}^{N} v_i^2}} = \dfrac{\langle u,v \rangle}{\lVert u \rVert \ \lVert v \rVert}
        \end{equation}
        \medskip
        
        where \textit{N} represents the number of dimensions composing the sentence embeddings \textit{u} and \textit{v}, $\langle u,v \rangle $ is the inner product between the two vectors, and $\lVert . \rVert $ is the L2 norm.
        
        % Ensemble models. Antecedentes. Ejemplos
        With the goal of building an accurate representation of each sentence, an ensemble approach has been adopted. The potential of this type of method to combine word embeddings has been assessed in the state-of-the-art literature~\cite{speer2019ensemble, yin2015learning}, showing that a mixture of embeddings featuring different characteristics leads to more robust representations and better performance than single embedding-based methods. Besides, a further advantage of ensemble methods is the expansion of vocabulary coverage.

        In the ensemble proposed, the output is calculated by concatenating the embeddings of four well-known multilingual models available at Sentence-Transformers\footnote{\url{https://www.sbert.net/}}~\cite{reimers_sentence-bert:_2019}, all of them fine-tuned on MSTSB\footnote{\url{https://github.com/Huertas97/Multilingual-STSB}}, a multilingual extended version of the Semantic Textual Similarity Benchmark (STSb)~\cite{cer-etal-2017-semeval}. Typically, Semantic Textual Similarity (STS) tasks include examples composed of a pair of sentences and a score ranging between 0 and
        5 according to the degree of similarity. STS Benchmark\footnote{http://ixa2.si.ehu.eus/stswiki/index.php/STSbenchmark} comprises a selection of the English datasets used in the STS tasks between 2012 and 2017 from SemEval. In order to work on a multilingual scenario, in this work, the STS Benchmark was extended to 15 different languages using the Google translator library. 
        
        %Ensemble methods are techniques that combine several models to produce improved results. Combining models by stacking or bagging are among the most well-known and widely used techniques in Machine Learning field \cite{bishop_pattern_2016,xia-ensemble}. 
        
        % However, as shown in \cite{yin2015learning}, it does not mean the more models, the better. Whether including a model to the set helps depends on the complementarity among the models. Consequently, exploring different combinations is needed.
        
        %\medskip
        
        \begin{Figure}
            \centering
            \includegraphics[width=1 \linewidth]{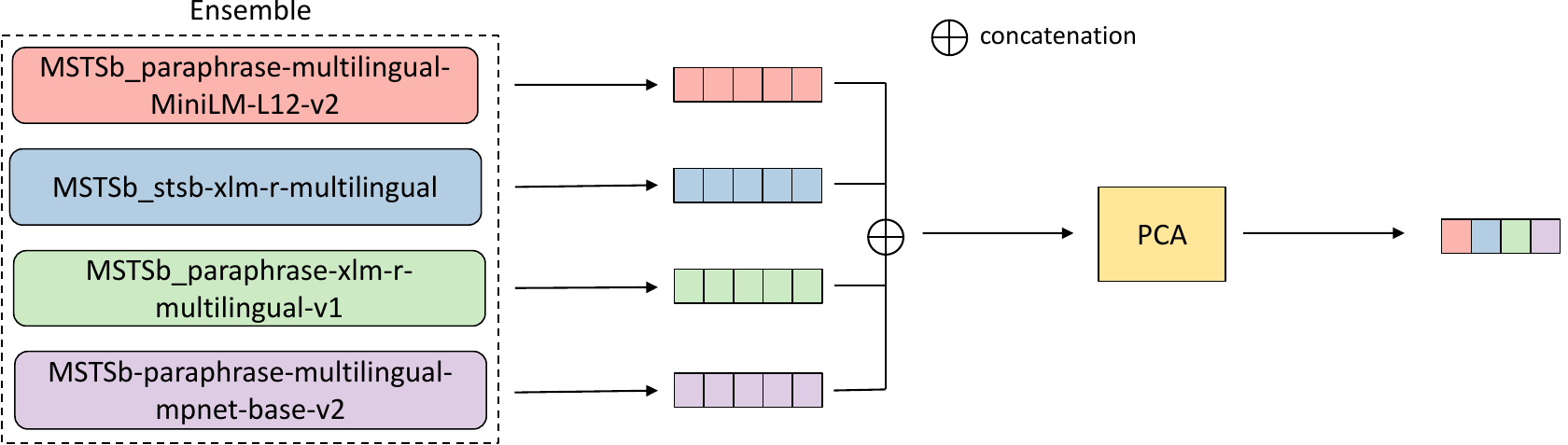}
            \captionof{figure}{Ensemble and dimensionality reduction approach proposed. Concatenation of embeddings from four multilingual sentence-transformers models applying PCA dimensionality reduction.}
            \label{Fig:Ensemble-diagram}
        \end{Figure}
        
        %\medskip
          
        The multilingual SentenceTransformers models used as base models in this study are:
          
        \begin{itemize}
            \item \textbf{paraphrase-xlm-r-multilingual-v1}: Distilled version of RoBERTa~\cite{liu_roberta:_2019} trained on large-scale paraphrase data using XLM-R~\cite{Conneau2019Nov} as the student model.
            
            \item \textbf{stsb-xlm-r-multilingual}: Distilled BERT~\cite{devlin2018bert} version trained in NLI~\cite{williams-etal-2018-broad} and STSb~\cite{cer-etal-2017-semeval} using XLM-R as the student model. 
            
            \item \textbf{paraphrase-multilingual-MiniLM-L12-v2}: Multilingual version of the MiniLM model from Microsoft~\cite{wang2020minilm} trained on large-scale paraphrase data. 
            
            \item \textbf{paraphrase-multilingual-mpnet-base-v2}: Distilled version of the MPNet model from Microsoft~\cite{song2020mpnet} fine-tuned with large-scale paraphrase data using XLM-R as the student model.
        \end{itemize}
        
        These pre-trained models are fine-tuned on MSTSB using Cosine Similarity Loss from Sentence Transformers~\cite{reimers_sentence-bert:_2019}. To obtain the best results and avoid overfitting, we optimized the following hyperparameters using the grid search method: learning rate, epochs, batch size, scheduler, and weight decay. The selected hyperparameter values and the resulting model have been published at HuggingFace\footnote{Fine-tuned models available in \href{https://huggingface.co/AIDA-UPM}{https://huggingface.co/AIDA-UPM}}.
         
        % Table of results ensemble and single models and PCA

        % Reducción de dimensionalidad importancia. PCA beneficios. Destacar como se selecciona para que se amultilingual y basada en similitud semántica
        As explained by Sidorov et al.~\cite{sidorov_soft_2014}, cosine similarity applied to a pair of \textit{N}-dimensional vectors has both time and memory \textit{O}(\textit{N}) complexity. That is, time and memory grow linearly with the number of dimensions of the vectors under comparison. This is the main drawback of the use of ensemble models on semantic search with sentence embedding. To address this issue, the Principle Component Analysis (PCA) is computed and applied to the whole architecture as shown in Fig.~\ref{Fig:Ensemble-diagram}. This enables to reduce dimensionality, removing redundant information across embeddings while retaining the most relevant information in a new \textit{N}-dimensional space.
        
        In order to maximise efficiency, the embedding of each fact-checked claim is precalculated. When receiving a new fact-checked claim, its embedding representation will be obtained applying the models of the ensemble and the PCA to the concatenated outputs and saved into the fact-checked claims database. This will allow to easily evaluate new claims, calculating the cosine distance to each fact-checked claim stored.

    \subsection{Natural Language Inference}
        \label{section:nli_description}
        Once a top-k corpus of hoaxes above a specific degree of semantic similarity has been identified, Natural Language Inference is used to infer the relation between the new input statement (hypothesis) and each fact-checked claim (premise). This relation may be \textit{entailment}, \textit{contradiction} or \textit{neutral}. While semantic similarity is unable to detect these finer nuances, a NLI model is able to detect a entailment or contradiction relationship given a pair of sentences. If we manage to detect if an statement entails a hoax, we can safely assume that the statement supports the hoax and therefore contains misinformation. Nevertheless, it is important to mention that Language Inference is not aware of the intentionality behind an statement, an issue which is not addressed in this research.
        
        To better describe the NLI task, let $\langle p, h \rangle$ be a sentence pair of hoax and statement. Using language inference we can infer \textit{contradiction} and \textit{neutral} probabilities, however, our main focus is on finding the degree of \textit{entailment}. We formally want to find if $h$, our statement, is a hoax $h_f$ or we are unable to determine the nature of the statement $h_u$. Formally we want to approximate Eq.~\ref{eq_2}.
        
        \begin{equation}\label{eq_2}
            f(p, h) \approx P(p|h_f)
        \end{equation}
        
        where $p$ is a hoax or fact-checked claim verified by fact-checkers and we have certainty that involves fake information, $h$ is the verifiable statement found by semantic similarity and $h_f$ is the event in which the statement contains misinformation. Therefore, our purpose is to find a suitable function $f$ that is able to approximate this probability. Finding $P(p|h_f)$ is equivalent to finding the probability of the entailment of $\langle p, h \rangle$. On the other hand we can safely say that $1 - P(p|h_f) = P(p|h_u)$ as the contradiction and neutrality of $\langle p, h \rangle$ does not give a meaningful explanation for $h$.

        In order to find $f$, the transformer model XLM-RoBERTa-large~\cite{Conneau2019Nov} is chosen. Transformer models for NLI have problems when transferring to unseen domains~\cite{Talman_Chatzikyriakidis_2019}, so special consideration is given to the fine-tuning process. To train this network, two datasets are used, XNLI~\cite{conneau2018xnli} and SICK~\cite{Marelli_Menini_Baroni_Bentivogli_Bernardi_Zamparelli_2014}. The inner transformer model XLM-R is fine-tuned first on XNLI. In this case, we used the model available at the Huggingface transformers repository\footnote{https://huggingface.co/joeddav/xlm-roberta-large-xnli}. After this step, a classification head is added to the model, which includes \textit{a)} a global average pooling of the last hidden state of the transformer model, \textit{b)} a linear layer with $768$ neurons and \textit{tanh} activation, \textit{c)} a $10\%$ dropout for training and \textit{d)} a classifier linear layer with \textit{softmax}. This classification head is trained on the SICK dataset, freezing the XLM-R weights to preserve previous pre-training. This is optimized using Adam \cite{Kingma2014Dec} optimizer with $0.001$ learning rate. The best weights are decided on the validation subset of SICK.

    \subsection{Semi-automated (S-AFC) fact-checking through Natural Language Inference and Semantic Similarity}
    
        In this work, we propose a 2 steps process to perform semi-automated fact-checking (S-AFC). The semantic similarity and Natural Language Inference modules described in the two previous sections (\ref{section:semantic_similarity_description} and \ref{section:nli_description}) are the pillars of this S-AFC process. The first step allows to filter an entire database of fact-checked statements or hoaxes, retrieving those that present semantic similarities with the new input claim. As a result, an ordered list by the degree of similarity is obtained, and the top $k$ results are selected. Then, the NLI module allows to perform language inference between the input claim and each candidate hoax in the top-$k$ result. If a fact-checked claim is found to correlate the input claim with enough certainty, the new claim is labelled.
        
        This two-step process (see Fig.~\ref{fig:facter_check}) is highly useful for different purposes. In addition to a semi-automated fact-checking of new claims that need to be checked, the combination of semantic similarity and Natural Language Inference can be used to analyse the evolution and presence of a particular statement in a large amount of data. For instance, in an Online Social Network such as Twitter, it is possible to filter thousands of tweets seeking for those that endorse or reject the statement.

    \subsection{Automated tracking of hoaxes in Twitter}
        
        The massive volume of information present on social media platforms makes it unmanageable to track and monitor hoaxes' evolution manually. For this reason, we propose an automatic social media tracking method based on the generation of search queries composed of keywords and search operators. These keywords are employed to extract information, such as tweets or posts, related to a given claim from the API of an social network. All the data download will offer an extensive view to study the evolution of a piece of misinformation.
        
        The use of keywords is due to the limitations that the API of these OSN impose. While searching for a given statement will only deliver tweets or posts replicating almost exactly the original input claim, the use of keywords aims to increase this search space and to obtain a wider picture. The method used for automatic keyword extraction is adapted from KeyBERT~\cite{grootendorst2020keybert}. KeyBERT is a keyword extraction technique that uses semantic-aware Transformer-based models to compute word and tweet embeddings and cosine similarity to
        find the most semantically similar words to the tweet. Accordingly, the most similar words are the keywords that best describe the tweet meaning. 
        
        Our proposal, named \textbf{FactTeR-ChecKey}, uses our multilingual \texttt{MSTSb-paraphrase-multilingual-mpnet-base-v2} model as the semantic-aware model. To optimise the multilingual keyword extraction, stopwords are removed by detecting the language with CLD2\footnote{\href{https://pypi.org/project/pycld2/}{https://pypi.org/project/pycld2/}} and removing the appropriate stop words with the NLTK toolkit~\cite{WagnerWiebke2010SBEK}. Additionally, the \texttt{bert-spanish-cased-finetuned-ner} from Hugging Face is included as the Name Entity Recognition (NER) model for Spanish. This NER model is applied only in Spanish, so the keyword extraction tool remains multilingual.
        
        % The main limitation of the keywords extraction module lies in the need for specific models for each language, due to the complexity of this task. Thus, in order to consider multiple languages in this step, different models are required.
        
% \texttt textsc

% \input{tables/keywords_queries_examples}
\begin{figure}[htpb] 
	\centering
	\includegraphics[width=1\columnwidth]{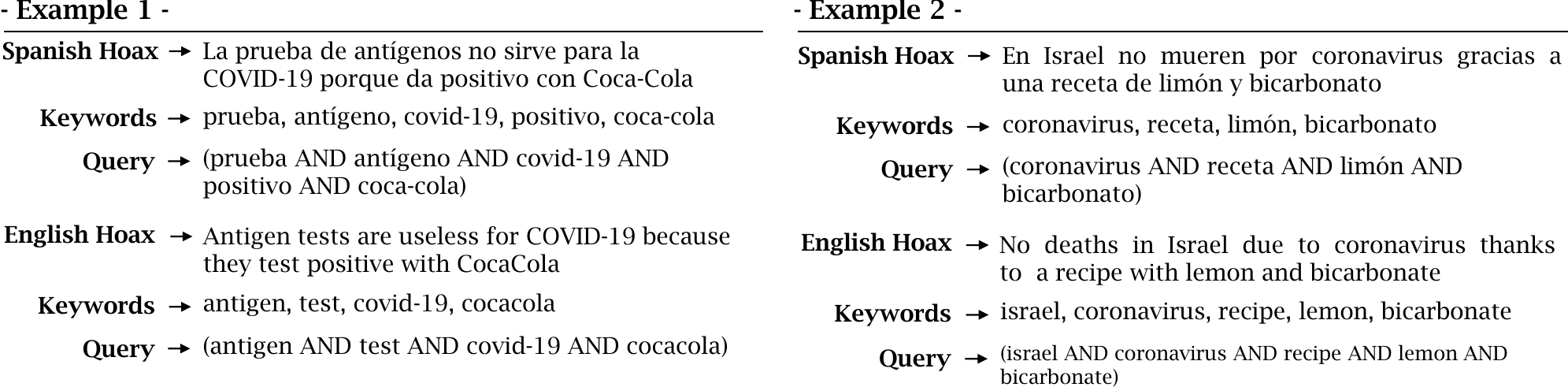}
\caption{Examples of query building from English and Spanish hoaxes for searching through Twitter API.} 
% The queries are built concatenating the different keywords extracted with the AND logical operator.}
\label{fig:keywords_example}
\end{figure}

\section{Evaluation of the FacTeR-Check architecture}
\label{section:evaluation}
    In this section, the Semantic Similarity, Natural Language Inference and keyword extraction (FactTeR-ChecKey) modules are evaluated using different benchmark datasets from the state-of-the-art literature. The following subsections describe in detail the results obtained for each task.

    \subsection{Semantic similarity evaluation}
            
        % PLOTS: Spearman correlation coefficient in fine-tuned models and ensembles with PCA
        \begin{figure}
            \centering
            \begin{subfigure}{0.45\linewidth}
                \includegraphics[width=\linewidth]{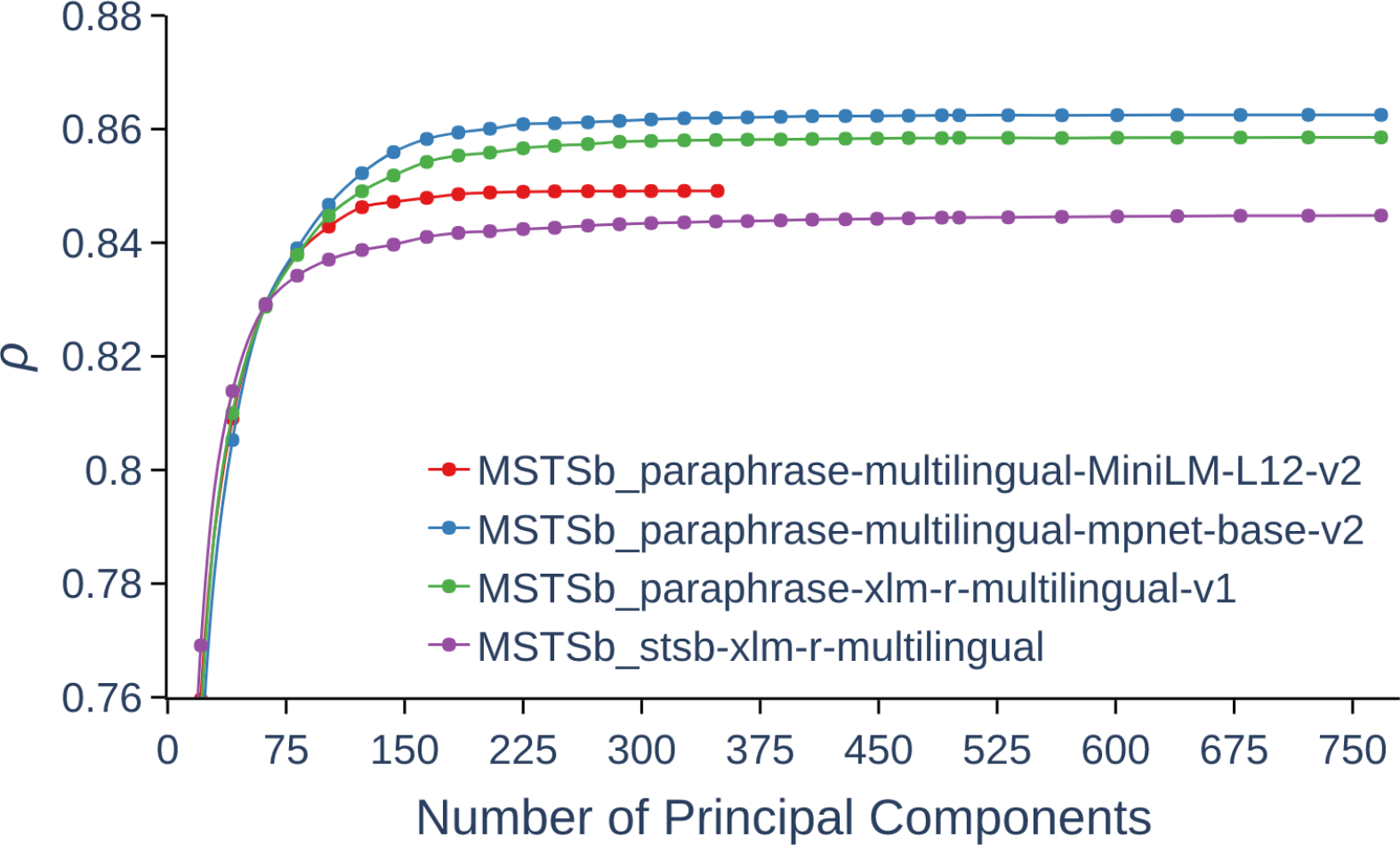}
                \caption{Single fine-tuned models}
                \label{fig:dev-pca-stsb:a}
            \end{subfigure}
            % \vfill 
            \hspace{4mm}
            \begin{subfigure}{0.5\linewidth}
                \includegraphics[width=\linewidth]{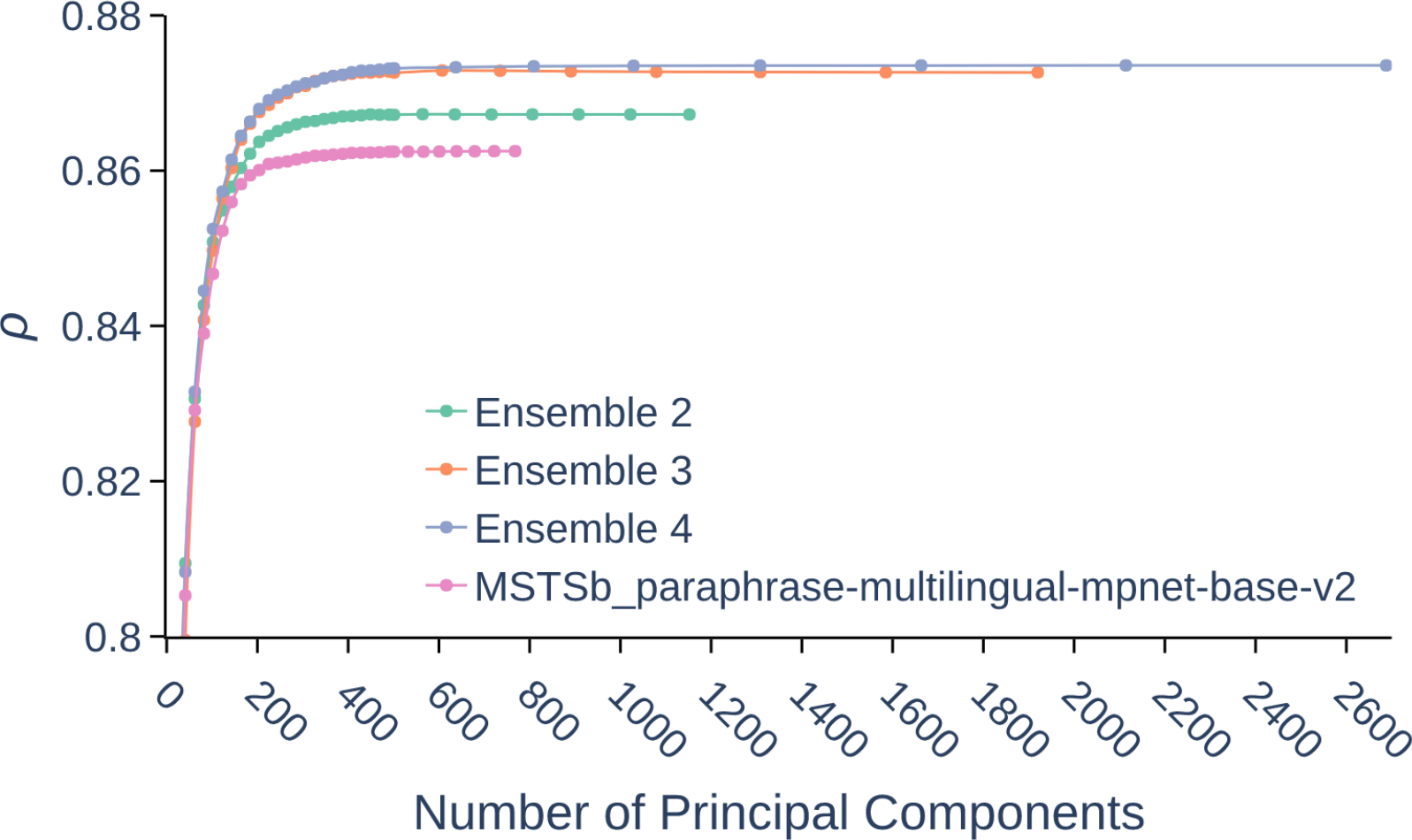}
                \caption{Ensemble architectures}
                \label{fig:dev-pca-stsb:b}
            \end{subfigure} 
            \medskip
            \caption{Number of components selection in MSTSB development set. Average Spearman Correlation Coefficient of the single fine-tuned models (a), and ensemble architectures (b) using cosine similarity for the 15 languages as a function of the number of components from the extended STS-Benchmark development set. The average of correlation coefficients is computed by transforming each correlation coefficient to a Fisher's z value, averaging them, and then back-transforming to a correlation coefficient.}
            \label{fig:dev-pca-stsb}
        \end{figure}

        % TABLE SPEARMAN STS
        \begin{table*}
            %\begin{subfigure}{\columnwidth}
            \centering
            \resizebox{\linewidth}{!}{%\resizebox{\columnwidth}{!}{%
            \begin{tabular}{lccccccccc} 
                \toprule
                \multicolumn{1}{c}{\multirow{2}{*}{\textbf{Model}}}   & \multirow{2}{*}{\textbf{Dimensions}} & \multicolumn{2}{c}{\textbf{EN-EN}} & \multicolumn{2}{c}{\textbf{EN-ES}} & \multicolumn{2}{c}{\textbf{ES-ES}} & \multicolumn{2}{c}{\textbf{Avg}}  \\
                \multicolumn{1}{c}{}                         &                             & $r$   & $\rho$            & $r$   & $\rho$            & $r$   & $\rho$            & $r$   & $\rho$           \\ 
                \midrule
                MSTSb\_paraphrase-multilingual-MiniLM-L12-v2 & 348                         & 85.26 & 86.17             & 81.45 & 81.49             & 83.30 & 83.68             & 81.38 & 81.47            \\
                MSTSb\_stsb-xlm-r-multilingual               & 768                         & 84.21 & 85.10             & 82.65 & 83.04             & 83.20 & 83.83             & 81.75 & 82.09            \\
                MSTSb\_paraphrase-xlm-r-multilingual-v1      & 768                         & 84.80 & 85.59             & 82.90 & 83.19             & 83.41 & 83.71             & 82.39 & 82.60            \\
                MSTSb-paraphrase-multilingual-mpnet-base-v2  & 768                         & 86.80 & \textbf{87.40}    & 84.42 & 84.45             & 85.19 & \textbf{85.52}    & 83.48 & 83.59            \\
                Ensemble 2                                   & 1152                        & 85.90 & 86.72             & 83.68 & 83.87             & 84.39 & 84.67             & 83.25 & 83.41            \\
                Ensemble 3                                   & 1920                        & 86.34 & 87.13             & 84.18 & 84.34             & 84.86 & 85.14             & 83.67 & \textbf{83.84 }  \\
                Ensemble 4                                   & 2688                        & 85.73 & 86.59             & 84.16 & \textbf{84.53}    & 84.67 & 85.25             & 83.33 & 83.62            \\
                \bottomrule
            \end{tabular}
            }
            
            \caption{Spearman $\rho$ and Pearson $r$ correlation coefficient between the sentence representation from multilingual models and the gold labels for STS Benchmark test set.}
            \label{table:STS_single_models_test}
        \end{table*}
            
        % TABLE SPEARMAN STS PCA    
        \begin{table*}
            %\begin{subfigure}{\columnwidth}
            \centering
            \resizebox{\linewidth}{!}{%\resizebox{\columnwidth}{!}{%
            
            \begin{tabular}{lccccccccc} 
                \toprule
                \multicolumn{1}{c}{\multirow{2}{*}{\textbf{Model + PCA}}} & \multirow{2}{*}{\textbf{Dimensions}} & \multicolumn{2}{c}{\textbf{EN-EN}} & \multicolumn{2}{c}{\textbf{EN-ES}} & \multicolumn{2}{c}{\textbf{ES-ES}} & \multicolumn{2}{c}{\textbf{Avg}}  \\
                \multicolumn{1}{c}{}                            &                             & $r$   & $\rho$            & $r$   & $\rho$            & $r$   & $\rho$            & $r$   & $\rho$           \\ 
                \midrule
                MSTSb\_paraphrase-multilingual-MiniLM-L12-v2    & 184                         & 84.92 & 85.71             & 81.04 & 81.04             & 83.08 & 83.28             & 81.03 & 81.02            \\
                MSTSb\_stsb-xlm-r-multilingual                  & 408                         & 84.35 & 85.11             & 82.84 & 83.17             & 83.39 & 83.89             & 81.85 & 82.08            \\
                MSTSb\_paraphrase-xlm-r-multilingual-v1         & 286                         & 84.79 & 85.50             & 82.73 & 82.97             & 83.38 & 83.58             & 82.23 & 82.39            \\
                MSTSb-paraphrase-multilingual-mpnet-base-v2     & 306                         & 86.69 & 87.27             & 84.21 & 84.28             & 84.93 & 85.19             & 83.20 & 83.28            \\
                Ensemble 2                                      & 347                         & 85.91 & 86.72             & 83.49 & 83.69             & 84.42 & 84.68             & 83.12 & 83.28            \\
                Ensemble 3                                      & 367                         & 86.64 & 87.55             & 84.50 & 84.80             & 85.24 & 85.72             & 83.85 & 84.21            \\
                Ensemble 4                                      & 429                         & 86.77 & \textbf{87.78}    & 85.00 & \textbf{85.52}    & 85.56 & \textbf{86.20}    & 84.24 & \textbf{84.71}   \\
                \bottomrule
            \end{tabular}
            }
            
            \caption{Spearman $\rho$ and Pearson $r$ correlation coefficient between the sentence representation from multilingual models with PCA dimenisonality reduction and the gold labels for STS Benchmark test set.}
            \label{table:STS_pca_models_test}
        \end{table*}

        The multilingual STS Benchmark (generated with Google Translator) has been used for the evaluation of the semantic similarity module. The overall results in the test sets are shown in Table~\ref{table:STS_single_models_test}. While the EN-EN column refers to the original STS Benchmark dataset, EN-ES, ES-ES are calculated using the translated version. These results reveal that the best performance is obtained with the fine-tuned \texttt{MSTSb-paraphrase-multilingual-mpnet-base-v2} model. This table also presents the results obtained with different combinations of the models. The best Ensemble of only 2 models is composed of the concatenation of \texttt{MSTSb\_paraphrase-xlm-r-multilingual-v1} and \texttt{MSTSb\_paraphrase-multilingual-MiniLM-L12-v2}, Ensemble 3 adds \texttt{MSTSb-paraphrase-multilingual-mpnet-base-v2} model while and Ensemble 4 includes all models reaching a maximum of 2688 dimensions. Surprisingly, only Ensemble 3 exceeds the best-fit model at the cost of incorporating more than twice as many dimensions. 
        
        As expected, the use of ensemble based approaches increases dramatically the number of dimensions. In order to tackle this problem, Principal Component Analysis (PCA) is used to reduce dimensionality. PCA is a data transformation and dimensionality reduction method that finds a subspace that explains most of the data variance while keeping attractive properties, such as removing linear correlation between dimensions and avoiding irrelevant dimensions with low variance. On the other hand, PCA is an unsupervised method that does not guarantee that the new feature space will be the most appropriate for a supervised task. To cope with this disadvantage, a total of 90K parallel sentences representing 15 languages\footnote{The languages used in this scenario are: ar, cs, de, en, es, fr, hi, it, ja, nl, pl, pt, ru, tr, zh} and extracted from three well-known resources (TED2020\footnote{https://www.ted.com/participate/translate}, WikiMatrix~\cite{schwenk_wikimatrix:_2019} and OPUS-NewsCommentary~\cite{tiedemann-2012-parallel}) are used to fit the PCA for each model. The relation between performance obtained and reduction size is shown in Fig.~\ref{fig:dev-pca-stsb}. As can be seen, both in the case of single fine-tuned models and ensemble architectures, the performance converges with less than 200 principal components, which provides a substantial space reduction. The best PCA space is selected according to the MSTSB development set average performance across languages.

        %% Resultados (brevemente vemos que el ensemble es el mejor y que aplicando PCA no perdemos mucho

        %   Los resultados de la Tabla  \ref{table:STS_pca_models_test} muestran como el mejor resultado se obtiene con el modelo fine-tuneado XXX. El Ensemble 2 está compuesto por la concatenación de XXX y XXX, el Ensemble 3 añade el modelo XXX y el Ensemble 4 incluye todos los modelos alcanzando un máximo de 2688 dimensiones. Sorpredentemente, sólo el ensemble 3 es capaz de mejorar en cierta medida al mejor modelo ajustado pero a costa de incorporar más del doble de dimensiones.  

        Table~\ref{table:STS_pca_models_test} shows the results after combining PCA and the ensemble approach, proving that this dimensionality reduction methods leads to better performance, reducing dramatically the number of dimensions. An illuminative example is Ensemble 4, which reduces from 2688 to 429 dimensions after applying PCA with the highest scores across all languages. This method not only reduces up to six times the initial dimensions of the ensemble, but it also requires fewer dimensions than most of the single models. This demonstrates that ensemble approaches in combination with dimensionality reduction techniques allow to build accurate and efficient semantic textual similarity models.
        
        %Overall, we have introduced a new multilingual semantic similarity approach based on the ensemble architecture with PCA dimensionality reduction evaluated in a multilingual scenario. 

    \subsection{Performance of the Natural Language Inference module}
    
        The NLI module is in charge of determining the relation between two statements (a fact-checked statement) and a new input claim. This relation, which can be either \textit{entailment}, \textit{contradiction} or \textit{neutral}, will be based on different probabilities. Thus, a threshold has to be defined in order to assign the final label. The most likely scenario is one with a large database of fact-checked claims verified by fact-checkers. Once a new claim has to be checked, it will be compared with the NLI module against those verified claims existing in the database above a certain degree of semantic similarity. As result, if enough degree of entailment is found, the new input claim will be labelled according to the verified claim found.
        
        We evaluate our approach using the testing subset provided by SICK. A well-known collection of pairs of sentences with \textit{entailment}, \textit{contradiction} and \textit{neutral} relation. Results are presented on Table~\ref{tab:sick_v2}. For comparison, we include the results of two benchmark methods: GenSen~\cite{subramanian2018learning} and InferSent~\cite{conneau2017supervised}. In case of GenSen, it achieves 87.8\% accuracy while InferSent reaches 0.863. Our proposed approach reaches 87.7\% accuracy while maintaining the multi-lingual capabilities of XLM-RoBERTa, which is useful to contrast information from culturally separated hoaxes. This is represented in the Spanish and interlingual sections of Table~\ref{tab:sick_v2}, where the same metrics are computed. We observe a slight drop in quality, mostly due to SICK being mono-lingual, though Spanish and inter-lingual results are quite robust on their own with 82.9\% and 85.3\% accuracy respectively. We want to highlight the high accuracy attained by the module when mixing languages, allowing for international tracking of misinformation.
        
        \begin{table*}[]
\begin{tabularx}{\textwidth}{lrrYYYY}

%\begin{tabular}{lrrllll}
\hline
\textbf{Language} &
  \multicolumn{1}{l}{} &
  \multicolumn{1}{l}{\textbf{}} &
  \textbf{Precision} &
  \textbf{Recall} &
  \textbf{F1-score} &
  \textbf{Support} \\ \hline
english & \multirow{3}{*}{Label}         & \multicolumn{1}{r|}{\textit{CONTRADICTION}} & 0.9158 & 0.7486 & 0.8238 & 712  \\
        &                                & \multicolumn{1}{r|}{\textit{ENTAILMENT}}    & 0.8475 & 0.8946 & 0.8704 & 1404 \\
        &                                & \multicolumn{1}{r|}{\textit{NEUTRAL}}       & 0.8856 & 0.9022 & 0.8938 & 2790 \\ \cline{2-7} 
        & \multirow{2}{*}{Summary}       & \multicolumn{1}{r|}{\textit{Macro Avg.}}    & 0.8830 & 0.8484 & 0.8627 & 4906 \\
        &                                & \multicolumn{1}{r|}{\textit{Weighted Avg.}} & 0.8791 & 0.8777 & 0.8770 & 4906 \\ \cline{2-7} 
 &
  \multicolumn{1}{c}{\textit{-}} &
  \multicolumn{1}{r|}{\textit{Accuracy}} &
  0.8777 &
  \multicolumn{1}{c}{-} &
  \multicolumn{1}{c}{-} &
  4906 \\ \hline
spanish & \multirow{3}{*}{Label}         & \multicolumn{1}{r|}{\textit{CONTRADICTION}} & 0.8511 & 0.7388 & 0.7910 & 712  \\
        &                                & \multicolumn{1}{r|}{\textit{ENTAILMENT}}    & 0.7446 & 0.9031 & 0.8162 & 1404 \\
        &                                & \multicolumn{1}{r|}{\textit{NEUTRAL}}       & 0.8797 & 0.8451 & 0.8461 & 2790 \\ \cline{2-7} 
        & \multirow{2}{*}{Summary}       & \multicolumn{1}{r|}{\textit{Macro Avg.}}    & 0.8251 & 0.8190 & 0.8178 & 4906 \\
        &                                & \multicolumn{1}{r|}{\textit{Weighted Avg.}} & 0.8369 & 0.8292 & 0.8296 & 4906 \\ \cline{2-7} 
        & \multicolumn{1}{c}{\textit{-}} & \multicolumn{1}{r|}{\textit{Accuracy}}      & 0.8292 & -      & -      & 4906 \\ \hline
\begin{tabular}[c]{@{}l@{}}inter\end{tabular} &
  \multirow{3}{*}{Label} &
  \multicolumn{1}{r|}{\textit{CONTRADICTION}} &
  0.8825 &
  0.8737 &
  0.8072 &
  1424 \\
        &                                & \multicolumn{1}{r|}{\textit{ENTAILMENT}}    & 0.7925 & 0.8989 & 0.8423 & 2808 \\
        &                                & \multicolumn{1}{r|}{\textit{NEUTRAL}}       & 0.8828 & 0.8586 & 0.8705 & 5580 \\ \cline{2-7} 
        & \multirow{2}{*}{Summary}       & \multicolumn{1}{r|}{\textit{Macro Avg.}}    & 0.8526 & 0.8337 & 0.84   & 9812 \\
        &                                & \multicolumn{1}{r|}{\textit{Weighted Avg.}} & 0.8569 & 0.8534 & 0.8533 & 9812 \\ \cline{2-7} 
        & \multicolumn{1}{c}{\textit{-}} & \multicolumn{1}{r|}{\textit{Accuracy}}      & 0.8534 & -      &  & 9812 \\ \hline
\end{tabularx}%

\caption{Results for the SICK test set. Spanish results are extracted from machine translations of the SICK test set. Interlingual results are made from pairing interchangeably Spanish and English prompts.}
\label{tab:sick_v2}
\end{table*}

    \subsection{Performance of the keywords extraction module}

        In order to evaluate the benefits of FactTeR-ChecKey, our approach is compared against two baseline methods in a general and a Twitter-specific scenario. The two baseline methods selected for this comparison are the statistical Rapid Automatic Keyword Extraction (RAKE) algorithm~\cite{RoseStuart2010AKEf} and the multilingual version of KeyBERT which use \texttt{paraphrase-xlm-r-multilingual-v1} as semantic-aware model. RAKE is a well-known statistical method for keyword extraction based on the collocation and co-occurrence of words by eliminating stopwords and punctuation, not taking into account any semantic information for the extraction process. On the other hand, KeyBERT incorporates state-of-the-art Transformer models for keyword extraction. The evaluation task consists on extracting keywords from the 60 Spanish hoaxes used previously in this project. Figure~\ref{fig:keywords_example} provides an overview of the hoaxes data and the queries built for searching through the Twitter API. The queries are built concatenating the different keywords extracted with the \textquote{AND} logical operator. Precision, recall, and F1 score are the metrics used to evaluate the ability to extract keywords compared to manually extracted keywords. 
        
        Due to the differences between a general search engine and a the Twitter search API\footnote{\href{https://twitter.com/search-advanced?lang=en}{https://twitter.com/search-advanced?lang=en}}, which entails several restrictions, we have evaluated the performance of FacTeR-ChecKey in both. While a common search engine such as Google allows rich queries and provides flexibility when using verbs as input, the Twitter search API is very restricted and only searches for the exact words used in the input. In the first stage of the project, in which hoax-related information was extracted with manually extracted keywords, it was observed that verbs limited the information retrieved due to these limitations. Therefore, verbs were removed from the Spanish keywords extracted for the Twitter scenario and an additional POS tagging filter was applied to the automatic keyword extracted. The POS tagging filter is performed using Spacy~\cite{spacy}, and the best model is selected from three possible models: small, medium, and large. It is noteworthy to highlight that although the automatic keyword extraction method is only evaluated on Spanish hoaxes, it can be easily extended to other languages. 
       
        \begin{table}[htpb]
   \centering
        \resizebox{0.55\linewidth}{!}{%
              \centering
                \begin{tabular}{lccc} 
                \toprule
                \multicolumn{1}{c}{\multirow{2}{*}{Keyword Model}}  & \multicolumn{3}{c}{General}                          \\
                                                & Precision       & Recall          & F1-score         \\ 
                \hline
                RAKE Spanish                    & 0.4683          & 0.9503          & 0.6139           \\
                KeyBERT Multilingual            & 0.5694          & 0.7121          & 0.6200           \\
                \textbf{FactTeR-ChecKey Multilingual}  & \textbf{0.6283} & \textbf{0.8267} & \textbf{0.6988}  \\
                \bottomrule
                \end{tabular}
                }
                \vspace{2mm}
                \caption{Evaluation of the keywords extraction module in general purpose tasks.}
                \label{table:keys_general}
            \end{table}%
        
\begin{table}[htpb]
  \centering
\resizebox{0.68\linewidth}{!}{%
    \begin{tabular}{lccc} 
    \toprule
    \multicolumn{1}{c}{\multirow{2}{*}{Keyword Model}}       & \multicolumn{3}{c}{Twitter}                          \\
                                          & Precision       & Recall          & F1-score         \\ 
    \hline
    RAKE spanish + POS Tag Filter  small     & 0.5692          & 0.8851          & 0.6731           \\
    RAKE spanish + POS Tag Filter  medium     & 0.5703          & 0.8952          & 0.6794           \\
    RAKE spanish + POS Tag Filter  large     & 0.5707          & 0.8590          & 0.6653           \\
    \hline
    KeyBERT + POS Tag Filter small           & 0.6433          & 0.6572          & 0.6293           \\
    KeyBERT + POS Tag Filter medium           & 0.6442          & 0.6577          & 0.6317           \\
    KeyBERT + POS Tag Filter large           & 0.6483          & 0.6242          & 0.6107           \\
    \hline
    FactTeR-ChecKey + POS Tag Filter small          & 0.6436          & 0.7397          & 0.6657           \\
    FactTeR-ChecKey + POS Tag Filter medium          & 0.6852          & 0.7847          & 0.7104           \\
    \textbf{FactTeR-ChecKey + POS Tag Filter large} & \textbf{0.6888} & \textbf{0.7917} & \textbf{0.7149}  \\
    \bottomrule
    \end{tabular}
    }
    \vspace{2mm}
    \caption{Evaluation of the keywords extraction module in the Twitter API.}
    \label{table:keys_twitter}
\end{table}

        % Analizar resultados
        Our technique clearly has an advantage over RAKE and KeyBERT approaches both in general (see Table~\ref{table:keys_general}) and Twitter-specific scenarios (see Table~\ref{table:keys_twitter}), where verbs are not considered. One advantage of FactTeR-ChecKey is that the type of information retrieved can be regulated by building queries from more specific to more general. Specific queries include all extracted keywords and gradually become more general as the terms are iteratively excluded from the query based on the similarity score. For this reason, our method has many practical applications. From already checked hoaxes, it is possible to extract information related to other hoaxes and to evaluate the check-worthiness of new hoaxes. 
        
        % Mostrar algún ejemplo donde funcione bien. 

\section{NLI19-SP: A Spanish Natural Language Inference dataset of hoaxes in Twitter}
\label{section:nli19-sp}    
    
    One of the goals of this research has been to build a dataset of tweets spreading misinformation claims detected and verified by fact-checkers. We have selected Twitter as the target OSN due to its large number of users, the availability of an API and the intensive movement of both information, misinformation and disinformation. Besides, our dataset is focused on misinformation spread in Spanish. To build such dataset, we have followed a four-step process:
            
    \begin{enumerate}
        \item \textbf{Hoaxes collection:} We gathered a pool of 61 hoaxes identified by fact-checker organisations.
        \item \textbf{Search queries generation:} It is necessary to build representative queries with keywords to retrieve tweets to the hoaxes from Twitter API
        \item \textbf{Tweets retrieving:} By using FacTeR-ChecKey, we built a search query for each of the hoaxes in order to download tweets related to them from the Twitter search API.
        \item \textbf{Filtering by semantic similarity:} We apply the semantic similarity module to filter tweets semantically related to each hoax.
        \item \textbf{Natural Language Inference labelling:} The NLI module is applied to label the tweets according to their relation with the original hoax, detecting those that support or contradict the false claim.
    \end{enumerate}
    
    The result of applying this pipeline is a pool of semantically-similar tweets for each hoax labelled as \textit{entailment}, meaning that the tweet endorses the false claim, \textit{contradiction} or \textit{neutral}.
    
    For the extraction of false claims already identified by fact-checkers we used LatamChequea Coronavirus\footnote{\url{https://chequeado.com/latamcoronavirus/}}, a database of misinformation about COVID-19 detected by 35 Spanish-language organisations and coordinated by Chequeado, and based on the global database created by the International Fact-checking Network (IFCN). Among all the indicators in this database, the variable used for our purpose will be the title of each false post registered. Given that the NLI and semantic similarity modules require the false claim to be expressed as clearly as possible, redundant words such as "hoax" or "message" that refer to the hoax itself are discarded.
    
    The second step involves the generation of search queries for each hoax through the FacTeR-ChecKey module. These search queries are then used through the Twitter API to find posts that are sharing that type of disinformation. Each search query generated was later manually enhanced to retrieve the maximum number of tweets spreading that false information. Each resulting query is composed of potential keywords from that falsehood, linked by search operators and the use of parentheses to improve the results.
    
    Furthermore, each set of keywords was optimised by adding synonyms and similar expressions to catch different ways to express the same piece of false information, because a hoax does not have to be propagated with the same words in the social network. This enables the collection of variants of the same hoax from different Hispanic geographical areas and avoids the implementation of a biased search of tweets from a unique Hispanic country.
    
    %For the purpose of this research, the query is considered acceptable and, thus, valid for the automated extraction of tweets under three conditions: firstly, returning many results throughout time, to the extent of concluding that this type of misinformation is being spread among users; secondly, not finding much noise (unrelated tweets) apart from tweets that are actually addressing that hoax or myth, and finally, only having fact-checking tweets warning of that misinformation instead of the misinforming tweets per se. If a specific query follows these three rules, we assign it an ID.
    
    The third step defines the automated search on Twitter API by using the search queries generated. This search is limited to the time period between the 1st of January 2020 to the 14th of March 2021. Moreover, replying tweets matching the query have not been excluded, since they can also misinformation. The result of this process comprises 61 queries selected for the automated search from reported hoaxes and tweets collected through them thanks to Twitter API. Appendix I shows the hoaxes in Spanish but and the English translation.
    
    In the next step the dataset has been curated using the semantic similarity module to filter tweets that actually present semantic similarity with the identified hoax. Finally, the Natural Language Inference component is applied to label each tweet as \textit{entailment}, \textit{contradiction} or \textit{neutral} according to the relation with the original hoax statement as presented by the fact-checkers. In accordance with Twitter regulations and in order to guarantee users’ privacy, users and texts will not be published.

\begin{figure*}
    \centering
    \includegraphics[width=0.9\textwidth]{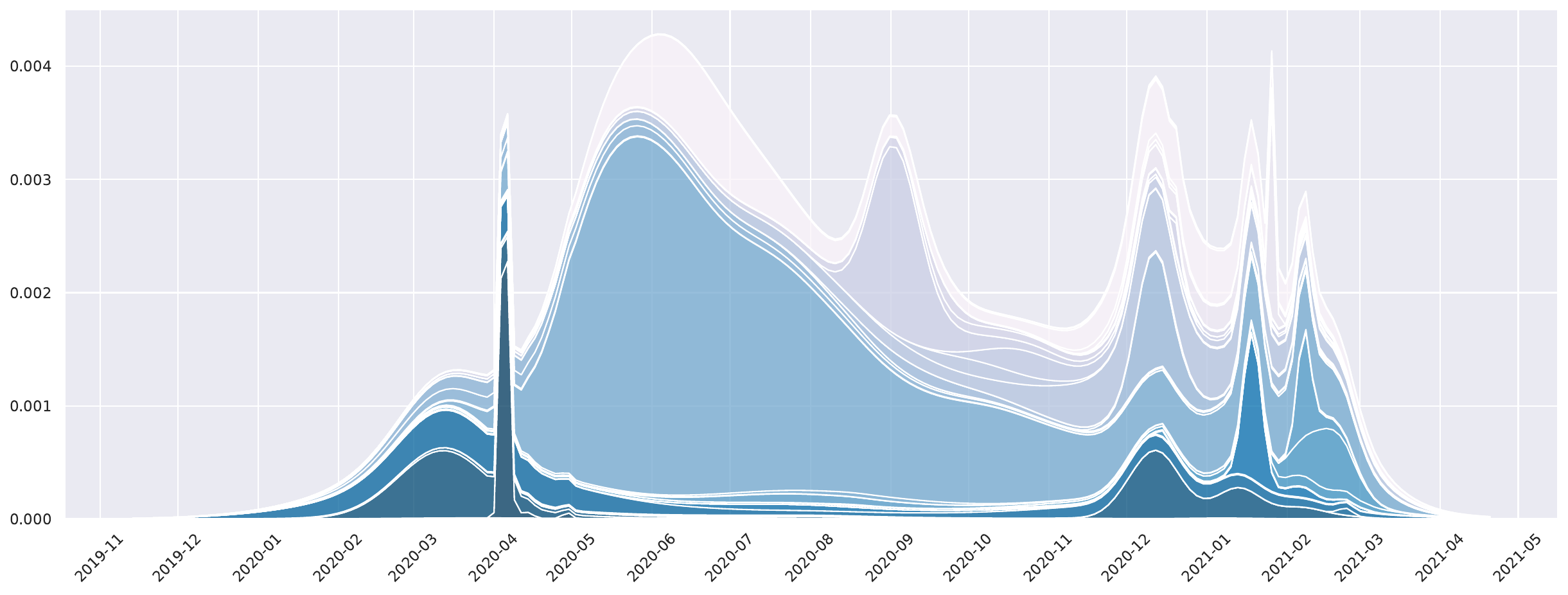}
    \captionof{figure}{Temporal distribution of tweets supporting the 61 hoaxes identified, evidencing common trends with multiple shared peaks.}
    \label{Fig:temporal_distribution_all_hoaxes}
\end{figure*}

\begin{figure*}
    \centering
    \includegraphics[width=0.9\textwidth]{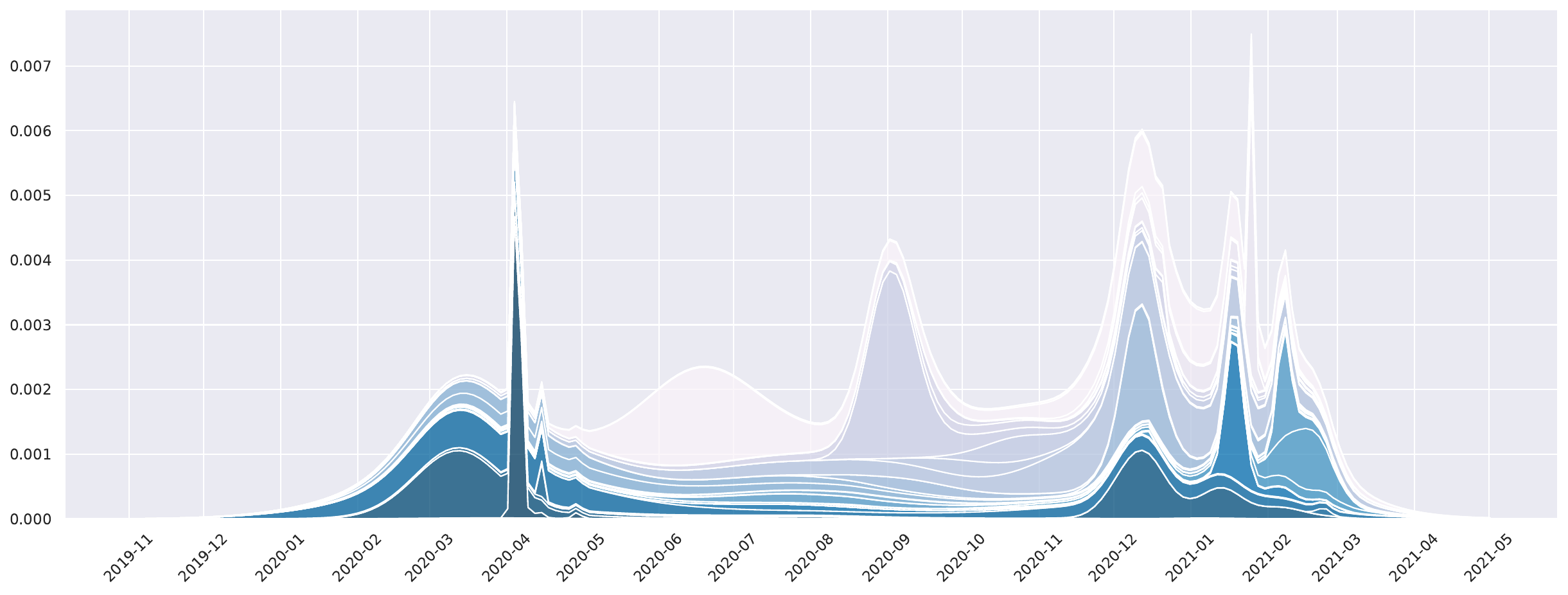}
    \captionof{figure}{Temporal distribution of tweets supporting the hoaxes identified without representing the hoax with id 31, related to the false claim ``masks cause hypoxia''.}
    \label{Fig:temporal_distribution_all_hoaxes_no_31}
\end{figure*}

\section{Misinformation spread in Spanish tweets: an analysis of Covid-19 hoaxes}
\label{section:analysis_twitter}
\begin{figure*}
    \centering
    \includegraphics[width=0.9\textwidth]{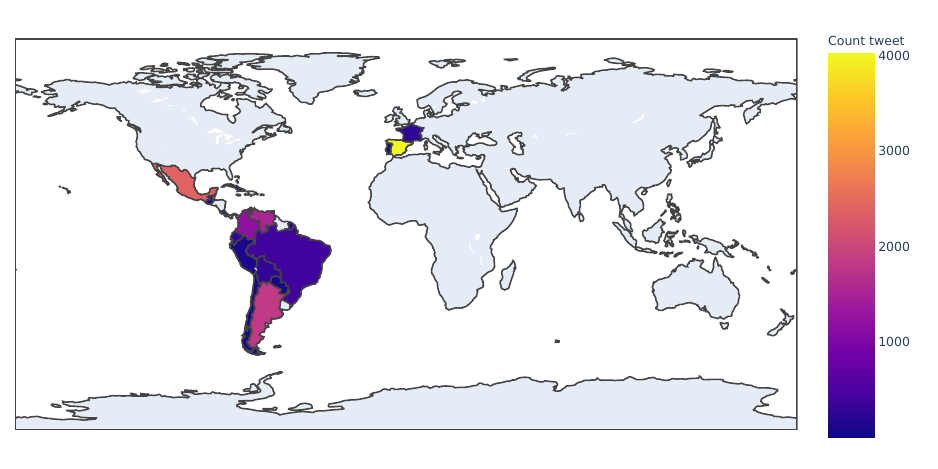}
    \captionof{figure}{Map showing the number of tweets supporting a hoax according to the nationality of the fact-checker that has identified it. In the case of France, although it is not a Spanish speaking country, several hoaxes have been identified by Factual AFP fact-checker, a France agency.}
    \label{Fig:countries_map}
\end{figure*}

In this section, our goal is to analyse how misinformation has spread in Twitter during the COVID-19 pandemic. For this purpose, we use the NLI19-SP dataset presented in the previous section. Each tweet in the dataset receives a label (entailment, contradiction or neutral) according to its relation with the most similar hoax. Additionally, tweets by Twitter accounts of fact-checkers have been also identified. All this information allows to infer relevant patterns and characteristics of misinformation and disinformation claims spread during the pandemic. To narrow the analysis, we focus on messages written in Spanish. Fig.~\ref{Fig:countries_map} shows the distribution of tweets found according to the fact-checker nationality that was used to identify the hoax. Although there is an important number of tweets collected from hoaxes identified by Spanish fact-checkers, no big differences were found between Spanish speaking countries.

Fig.~\ref{Fig:temporal_distribution_all_hoaxes} shows cumulative distribution plot for a general overview of the tweets collected that support the different hoaxes, represented with different colours. One of the most relevant conclusions that can be extracted from this analysis lies in the shared patterns among the different hoaxes, exhibiting a clear trend towards waves of misinformation. This behaviour reflects how misinformation inevitably feeds itself and how spreaders operate in a coordinated fashion, giving rise to waves of misinformation and disinformation. This phenomenon is also worth considering when taking steps to counter the propagation of misinformation. Besides, the large representation of specific hoaxes is also an important element to study. Thus, one of the most disseminated hoax (Hoax 31 in Table~\ref{tab:hoaxes_2}) is that ``masks cause hipoxia''. The large number of tweets found supporting this false claim is the reason of the big wave centred on June 2020. Similarly, the peak located at April 2020 y mainly due to the hoax ``Christine Lagarde said that the elderly live too long''.

In order to better visualise the distribution of tweets supporting hoaxes, in Fig.~\ref{Fig:temporal_distribution_all_hoaxes_no_31} the same plot is displayed without including the hoax 31, which concentrates large part of the tweets. Although the big wave disappears in this new plot, reflecting that it was caused by the hoax removed, one can see how the waves are still visible, evidencing the common behavioural patterns that describe how misinformation circulates. 

\begin{figure*}
    \centering
    \includegraphics[width=0.9\textwidth]{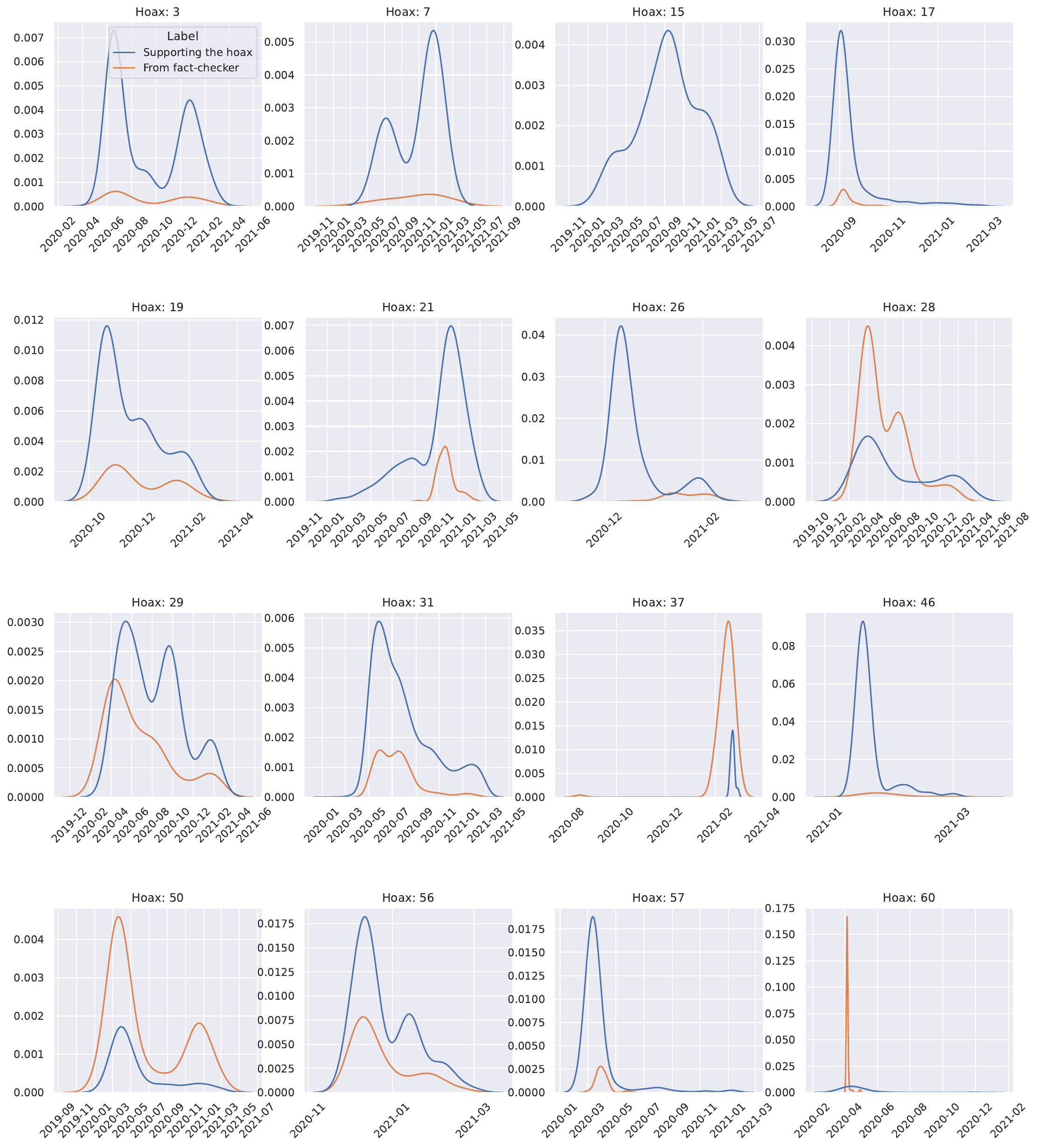}
    \captionof{figure}{Comparative for different hoaxes between the distribution of tweets supporting a specific hoax and tweet by fact-checkers rejecting it.}
    \label{Fig:histograms}
\end{figure*}

For a deeper analysis of misinformation circulating during the Covid-19 pandemic, Fig.~\ref{Fig:histograms} shows the temporal distribution of tweets supporting a selection of hoaxes and tweets published by fact-checker Twitter accounts. In four cases, hoaxes 28, 37, 50 and 60, the campaign launched by fact-checking organisation resulted in a higher number of tweets countering the hoax that tweets actually supporting the hoax. For the rest of hoaxes analysed, fact-checkers started a very timid response. However, in case of the hoax 15, no presence of fact-checkers denying the hoax can be appreciated, a false claim stating that ``The definition of pandemic was changed in 2009 by the WHO''. This reveals how complex is this scenario and that further research is required in order to help fact-checkers to detect and to undertake activities to avoid the spreading of false claims. In any case, it must be taken into consideration that the response must be proportionate, avoiding an excessive response that could increase the dissemination of the hoax and amplify its effects.

%   \begin{Figure}
%  \centering
%  \includegraphics[width=0.9 \linewidth]{figures/entailment_threshold_evolution.pdf}
%  \captionof{figure}{Evolution of the number of tweets labelled as entailment, neutral or contradiction according to the threshold applied.}
%  \label{Fig:entailment_evo}
% \end{Figure}

\section{Conclusion}
%Resumen de la propuesta
In this article we have proposed FacTeR-Check to mitigate OSN misinformation. Our architecture provides two pipelines, one for semi-automated verification of claims; another for tracking known hoaxes on social media. The pipelines share three modules: a semantic similarity module, a NLI module and a information retrieval module. By using context-aware semantic similarity, we are able to find related fact-checks, while NLI allows to contrast the claim against reputable sources. This double process enables to perform semi-automated fact-checking. On the other hand, in order to track hoaxes, we retrieve tweets related to a hoax, filtering the most relevant tweets with semantic similarity and contrasting them with the original hoax, finding how this particular piece of misinformation has spread on a social media platform. While our validation has been limited to COVID-19 and Twitter we want to emphasise that our architecture is adaptable to other knowledge domains as well as other social networks.

%Resumen de la evaluacion
For the evaluation, we first assess each model individually. Then the modules are put together in both pipelines to test their joint performance. To begin with, the similarity module offers above average performance using multilingual models on the STS benchmark. The NLI module uses XLM-RoBERTa fine-tuned on XNLI and the SICK training dataset, which performs adequately on SICK test, offering similar results to state-of-the-art models, as well as offering multilingual capabilities. Finally, the information retrieval module is compared against KeyBERT and RAKE on a dataset of Spanish keywords from our gathered hoaxes. 
%Insights
Using this architecture we built a dataset for misinformation detection using NLI in Spanish about COVID-19, as well as track a selection of hoaxes to analyse their spread. FacTeR-Check proves to extract insightful information about the spread of many hoaxes, showing aggregate frequency peaks matching COVID-19 waves in Spain. Identified hoaxes have their own particular activity peaks, some have more longevity than others, others are used much more; they are extremely diverse in lifetime and popularity.

%Future works
In contrat to previous approaches, FacTer-Check relies on external databases to operate. If a rumour reaches the verification pipeline, and there is no related fact-check retrievable on the topic, only similar articles will be retrieved. This means that the verification pipeline is as robust as the fact-check database. Alternatives may include composing a massive database of hoax embeddings, as well as a dynamic information retrieval process to detect new hoaxes and calculate their embeddings. The architecture has been tested on OSNs, meaning that it is blind to outside information such as news sites or other valuable sources of information. If a piece of disinformation is published outside of the OSN, it will be out of the scope of the tracking algorithm. Finally, information is varied, coming in many shapes and forms, including text but also audio, video or images; the verification and tracking pipeline can only work on textual data, meaning that there is room for building systems that support other formats.

\section*{Acknowledgement}

This work has been supported by the research project CIVIC: Intelligent characterisation of the veracity of the information related to COVID-19, granted by BBVA FOUNDATION GRANTS FOR SCIENTIFIC RESEARCH TEAMS SARS-CoV-2 and COVID-19, by the Spanish Ministry of Science and Innovation under FightDIS (PID2020-117263GB-100) and XAI-Disinfodemics (PLEC2021-007681) grants, by Comunidad Aut\'{o}noma de Madrid under S2018/TCS-4566 grant, by European Comission under IBERIFIER - Iberian Digital Media Research and Fact-Checking Hub (2020-EU-IA-0252), and by "Convenio Plurianual with the Universidad Politécnica de Madrid in the actuation line of \textit{Programa de Excelencia para el Profesorado Universitario}".

%Agradecemos el 

% \ifCLASSOPTIONcaptionsoff
%   \newpage
% \fi



\bibliographystyle{IEEEtran}
\bibliography{bibtex.bib}

\clearpage
\onecolumn
\appendix
\medskip
%\section{List of hoaxes analysed}
\label{appendix:list_hoaxes}

%\includegraphics[scale=0.9, angle=-90]{myTable.pdf}
% \usepackage{booktabs}
% \usepackage{graphicx}
\begin{table*}[htb]
\centering
\resizebox{1\textwidth}{!}{%
\begin{tabularx}{1.4\textwidth}{lXXX}

\toprule
\textbf{Id} & \textbf{Hoax (in Spanish)} & \textbf{Hoax (in English)} & \textbf{Fact-checkers} \\ \midrule
1 & La PCR no distingue entre coronavirus y gripe & PCR tests do not distinguish between coronavirus and the flu & Newtral.es \\ \midrule
2 & Las vacunas de ARN-m contra el coronavirus nos transforman en seres transgénicos & mRNA vaccines against coronavirus transform us into transgenic beings & Animal Político, Maldita.es, Newtral.es \\ \midrule
3 & La vacuna contra la COVID-19 se crea con células de fetos abortados & COVID-19 vaccines are made of cells from aborted fetuses & Agencia Ocote, Agência Lupa, Chequeado, ColombiaCheck, Maldita.es, Newtral.es \\ \midrule
4 & Merck asocia las vacunas contra la COVID-19 con un genocidio & Merck associates COVID-19 vaccines with a genocide & Ecuador Chequea, Newtral.es \\ \midrule
5 & Una imagen relaciona la prueba PCR con la destrucción de la glándula pineal en el Antiguo Egipto & An image links PCR tests to the destruction of the pineal gland & Maldita.es \\ \midrule
6 & La vacuna contra la COVID-19 produce parálisis facial & COVID-19 vaccines produce facial paralysis & Chequeado, Newtral.es \\ \midrule
7 & La primera ministra de Australia finge ponerse la vacuna contra la COVID-19 & Australia first minister pretends to get the COVID-19 vaccine & Agência Lupa, La Silla Vacía \\ \midrule
8 & La vacuna contra la COVID-19 produce convulsiones & COVID-19 vaccines produce seizures & Maldita.es, Newtral.es \\ \midrule
9 & Mueren 53 personas en Gibraltar tras ponerse la vacuna contra la COVID-19 & 53 people dead after being vaccinated against COVID-19 in Gibraltar & Maldita.es, Newtral.es \\ \midrule
10 & Detienen en un Lidl de Gijón a 11 personas con COVID-19 & 11 people with COVID-19 arrested in Lidl supermarket in Gijón & Maldita.es, Newtral.es \\ \midrule
11 & Ya no existen las enfermedades respiratorias que no son COVID-19 & Respiratory diseases that are not COVID-19 do not exist anymore & Newtral.es \\ \midrule
12 & La PCR da positivo por nuestros genes endógenos, no por coronavirus & PCR tests positive due to our endogenous genes, not due to coronavirus & Newtral.es \\ \midrule
13 & La ciudad de Rosario (Argentina) para la vacunación por los efectos adversos de la vacuna & The city of Rosario (Argentina) stops vaccination because of the adverse effects of the vaccine & Chequeado, Maldita.es \\ \midrule
14 & La OMS dice que llevar a los niños al colegio sirve como consentimiento para su vacunación & The WHO says that taking our children to school gives consent for their vaccination & Maldita.es \\ \midrule
15 & La definición de pandemia cambió en 2009 por la OMS & The definition of pandemic was changed in 2009 by the WHO & Newtral.es \\ \midrule
16 & Muere una enfermera de Tennessee (Estados Unidos) tras vacunarse contra la COVID-19 & A nurse from Tennessee (United States) died after being vaccinated against COVID-19 & La Silla Vacía, Maldita.es, Newtral.es \\ \midrule
17 & Solo el 6\% de las muertes por coronavirus en Estados Unidos fueron realmente por esta causa & Only 6\% of coronavirus deaths in United States were actually due to this cause & AFP, Agência Lupa, Animal Político, Chequeado, La Silla Vacía \\ \midrule
18 & La PCR da positivo por los exosomas, no por coronavirus & PCR tests positive due to exosomes, not due to coronavirus & Newtral.es \\ \midrule
19 & La mascarilla produce enfermedades neurodegenerativas & Masks produce neurodegenerative diseases & Maldita.es, Newtral.es \\ \midrule
20 & En Países Bajos existe desde 2015 una patente de test de COVID-19 & A patent of COVID-19 test exists in the Netherlands since 2015 & Maldita.es, Newtral.es \\ \midrule
21 & La vacuna contra la COVID-19 causa esterilidad & Pfizer vaccines cause sterility & Animal Político, Chequeado, ColombiaCheck, La Silla Vacía, Maldita.es, Newtral.es \\ \midrule
22 & Un estudio de 2008 financiado por la Comisión Europea ya incluía la COVID-19 & A study funded by the European Commission in 2008 already included COVID-19 & Newtral.es \\ \midrule
23 & Varios vacunados con la vacuna UQ-CSL contra la COVID-19 contraen el VIH & Several COVID-19 vaccinated people with UQ-CSL contracted HIV & Newtral.es \\ \midrule
24 & La vacuna contra la COVID-19 es aún experimental porque está en fase 4 & Vaccines against COVID-19 are still experimental because they are in phase 4 & Animal Político, Maldita.es \\ \midrule
25 & El Banco Mundial tenía planes para la COVID-19 desde 2017 & The World Bank had plans for COVID-19 since 2017 & Animal Político, Aos Fatos, Mala Espina Check \\ \midrule
26 & La vacuna contra la COVID-19 destruye nuestro sistema inmunológico & Vaccines against COVID-19 destroy our immune system & Maldita.es, Newtral.es \\ \midrule
27 & Pirbright Institute patentó la COVID-19 en 2018 & Pirbright Institute patented COVID-19 in 2018 & Maldita.es \\ \midrule
28 & Las gargaras con agua y sal previenen o curan el coronavirus & Gargling with water and salt prevents or cures coronavirus & \#NoComaCuento (La Nación), AFP, Chequeado, ColombiaCheck, Ecuador Chequea, Efecto Cocuyo, El Surtidor, La Silla Vacía, Maldita.es, Spondeo Media, Verificador (La República) \\ \midrule
29 & La dieta alcalina previene o cura el coronavirus & Alcaline diets prevent or cure coronavirus & Agência Lupa, Animal Político, Bolivia Verifica, Chequeado, ColombiaCheck, Cotejo.info, EFE Verifica, Ecuador Chequea, Efecto Cocuyo, \#NoComaCuento (La Nación), La Silla Vacía, Mala Espina Check, Maldita.es, Newtral.es \\ \midrule
30 & El coronavirus fue fabricado en un laboratorio chino & Coronavirus was made in a Chinese lab & Chequeado, Ecuador Chequea, Estadão verifica \\ \bottomrule
\end{tabularx}%
}
\caption{Relation of hoaxes 1 - 30.}
\label{tab:hoaxes_1}
\end{table*}

\begin{table*}[]
\centering
\resizebox{\textwidth}{!}{%
\begin{tabularx}{1.4\textwidth}{lXXX}

\toprule
\textbf{Id} & \textbf{Hoax (in Spanish)} & \textbf{Hoax (in English)} & \textbf{Fact-checkers} \\ \midrule
31 & La mascarilla causa hipoxia & Masks cause hypoxia & Agencia Ocote, Agência Lupa, Animal Político, Aos Fatos, Bolivia Verifica, Chequeado, ColombiaCheck, Cotejo.info, EFE Verifica, Ecuador Chequea, Efecto Cocuyo, La Silla Vacía, Maldita.es, Newtral.es, Verificado, Verificador (La República) \\ \midrule
32 & El eucalipto previene o cura el coronavirus & Eucalyptus prevents or cures coronavirus & AFP \\ \midrule
33 & El matico cura el coronavirus & Matico cures coronavirus & Bolivia Verifica \\ \midrule
34 & El biomagnetismo mata el coronavirus & Biomagnetism kills coronavirus & Bolivia Verifica, Maldita.es \\ \midrule
35 & La hoja de guayaba previene o cura el coronavirus & Guava leaf prevents or cures coronavirus & Animal Político, Bolivia Verifica, Maldita.es, Newtral.es \\ \midrule
36 & La NASA catalogó el dióxido de cloro como antídoto universal en 1988 & NASA catalogued chlorine dioxide as a universal antidote in 1988 & Animal Político \\ \midrule
37 & El vino previene o cura el coronavirus & Wine prevents or cures coronavirus & Chequeado, EFE Verifica, Maldita.es, Newtral.es \\ \midrule
38 & La mascarila causa la muerte por neumonía bacteriana & Masks cause death due to bacterial pneumonia & Maldita.es \\ \midrule
39 & La vitamina C previene o cura el coronavirus & Vitamin C prevents or cures coronavirus & AFP, Chequeado, EFE Verifica, Agência Lupa, Maldita.es, Verificado \\ \midrule
40 & La prueba de antígenos no sirve para la COVID-19 porque da positivo con Coca-Cola & Antigen tests are useless for COVID-19 because they test positive with Coca-cola & Maldita.es, Newtral.es \\ \midrule
41 & La homeopatía previene o cura el coronavirus & Homeopathy prevents or cures coronavirus & Chequeado, Mala Espina Check, Maldita.es, Periodismo de barrio / El Toque \\ \midrule
42 & La COVID-19, el MERS y el H1N1 coinciden con la instalación del 5G, 4G y 3G, respectivamente & COVID-19, MERS and H1N1 coincide with the installation of 3G, 4G and 5G, respectively & Poligrafo \\ \midrule
43 & Los indígenas protegen a los niños con hierbas y árboles frente a la COVID-19 & indigenous groups protect their children from COVID-19 with herbs and trees & Ecuador Chequea \\ \midrule
44 & Los mosquitos transmiten el coronavirus de contagiados & Mosquitoes transfer coronavirus from infected people & Maldita.es \\ \midrule
45 & Beber agua o sorbos previene o cura el coronavirus & Drinking or sipping water prevents or cures coronavirus & \#NoComaCuento (La Nación), AFP, Bolivia Verifica, ColombiaCheck, La Silla Vacía, Maldita.es, OjoPúblico \\ \midrule
46 & Mueren 55 personas en Estados Unidos tras vacunarse contra la COVID-19 & 55 people dead after being vaccinated against COVID-19 in the United States & EFE Verifica \\ \midrule
47 & Las mascarillas producen pleuresia y neumonía & Masks produce pneumonia and pleurisy & AFP \\ \midrule
48 & Las personas sanas llevan la mascarilla con la parte blanca hacia fuera para no contagiarse de COVID-19 & Healthy people wear their masks with the white part on the outside not to get COVID-19 & Newtral.es \\ \midrule
49 & El SARS-COV-2 no cumple los postulados de Koch, Rivers e Inglis para considerarlo enfermedad y coronavirus & SARS-COV-2 does not fulfill Koch, Rivers and Inglis' postulates in order to be considered as coronavirus and as a disease & EFE Verifica \\ \midrule
50 & Christine Lagarde dijo que los ancianos viven demasiado & Christine Lagarde said that the elderly live too long & Chequeado, ColombiaCheck, Ecuador Chequea, Maldita.es \\ \midrule
51 & La COVID-19 es una bacteria & COVID-19 is a bacteria & Animal Político, Chequeado, ColombiaCheck, La Silla Vacía, Maldita.es, Verificador (La República) \\ \midrule
52 & Galicia aprueba una ley para aislar a los positivos COVID-19 en campos de concentración & Galicia approves a law to aisle COVID-19 positives in concentration camps & Maldita.es \\ \midrule
53 & Las ondas electromagnéticas del 5G propagan el coronavirus & 5G electromagnetic waves spread coronavirus & Chequeado, Ecuador Chequea \\ \midrule
54 & La OMS rcomienda un test pulmonar para identificar el coronavirus & The WHO recommends a pulmonary test to detect coronavirus & EFE Verifica \\ \midrule
55 & Las pandemias tienen lugar cada 100 años & Pandemics take place every 100 years & AFP, Animal Político, ColombiaCheck, Verificador (La República) \\ \midrule
56 & El laboratorio de Wuhan tiene relación con Glaxo y Pfizer & Wuhan lab is related to Glaxo and Pfizer & Animal Político, Chequeado, La Silla Vacía, Maldita.es, Newtral.es \\ \midrule
57 & El coronavirus desaparece a los 27 grados & Coronavirus disappears at 27 degrees & Bolivia Verifica, Convoca, Agência Lupa \\ \midrule
58 & Hubo 17000 y 26000 muertes más en 2019 y 2018 respectivamente que en 2020 & There were 17000 and 26000 more deaths in 2019 and 2018 respectively than in 2020 & Maldita.es, Newtral.es \\ \midrule
59 & El polisorbato 80 de la vacuna contra la gripe causa coronavirus & Polysorbate 80 in the flu vaccines cause coronavirus & EFE Verifica, Maldita.es \\ \midrule
60 & Detienen a Charles Lieber por crear y vender el coronavirus & Charles Lieber arrested for creating and selling coronavirus & \#NoComaCuento (La Nación), AFP, Animal Político, Efecto Cocuyo, Agência Lupa, Mala Espina Check, Maldita.es, Newtral.es \\ \midrule
61 & En Israel no mueren por coronavirus gracias a una receta de limón y bicarbonato & No deaths in Israel due to coronavirus thanks to a recipe with lemon and bicarbonate & Newtral.es, Verificado \\ \bottomrule
\end{tabularx}%
}
\caption{Relation of hoaxes 31 - 61}
\label{tab:hoaxes_2}
\end{table*}

\end{document}